\documentclass{article}

\PassOptionsToPackage{numbers}{natbib}
\usepackage[eandd, final]{neurips_2026}
\usepackage[utf8]{inputenc} % allow utf-8 input
\usepackage[T1]{fontenc}    % use 8-bit T1 fonts
\usepackage{hyperref}       % hyperlinks
\hypersetup{hidelinks}
\usepackage{url}            % simple URL typesetting
\usepackage{booktabs}       % professional-quality tables
\usepackage{amsfonts}       % blackboard math symbols
\usepackage{nicefrac}       % compact symbols for 1/2, etc.
\usepackage{microtype}      % microtypography
\usepackage{xcolor}         % colors
\usepackage{tabularx}
\usepackage{graphicx}
\usepackage{amsmath}
\usepackage{multirow}
\usepackage{booktabs}
\usepackage{amssymb}
\usepackage{dsfont}
\usepackage{longtable}   
\usepackage{subcaption}
\usepackage{caption}
\usepackage{tabularx}
\usepackage{wrapfig}
\title{ColorConceptBench: A Benchmark for Probabilistic Color-Concept Understanding in Text-to-Image Models}

\usepackage{xspace}
\newcommand{\bench}{\emph{ColorConceptBench}\xspace}
\newcommand{\q}[1]{\emph{`#1'}}

\author{%
  Chenxi Ruan$^1$ \quad Yihan Hou$^1$\thanks{Corresponding author} \quad Yu Xiao$^1$  \quad
  Guosheng Hu$^3$ \quad Wei Zeng$^{1,2}$ \\
  $^1$The Hong Kong University of Science and Technology (Guangzhou), Guangzhou, China \\
  $^2$The Hong Kong University of Science and Technology, Hong Kong SAR, China \\
  $^3$China Academy of Art, Hangzhou, China \\
  \texttt{cruan361@connect.hkust-gz.edu.cn}\\
  Project: \href{https://huggingface.co/datasets/ColorConceptBench/ColorConceptBench}{ColorConceptBench/Dataset\&Code}
}

\begin{document}

\maketitle

\begin{abstract}
Text-to-image (T2I) models have advanced considerably in generating high-quality images from textual descriptions. 
However, their ability to associate colors with concepts remains largely constrained to explicit color names or codes, while their capacity to handle \emph{implicit concepts}, such as emotions and visual states, remains underexplored.
To address this gap, we introduce \bench, an expert-annotated benchmark that systematically evaluates color-concept associations through probabilistic color distributions.
\bench moves beyond explicit color specifications by examining how models interpret 1,281 implicit color concepts, grounded in 6,584 human annotations.
Our evaluation of nine leading T2I models reveals that performance varies substantially across semantic categories, and models exhibit a significant lack of sensitivity to abstract semantics.
These limitations persist even when applying classifier-free guidance scaling at inference time, suggesting that achieving human-like color understanding demands a shift in how models learn and represent implicit semantic meaning.
\end{abstract}
\section{Introduction}

Recent advances in text-to-image (T2I) generation can synthesize high-fidelity images that align semantically with text prompts,
demonstrating sophisticated control over object composition \citep{huang2023t2icompbench,ghosh2023geneval} and spatial relationships \citep{bakr2023hrs,gokhale2022benchmarking}.
Yet a significant challenge remains: these models often struggle to accurately understand and render color semantics, particularly in tasks requiring precise color rendering and complex attribute binding \citep{liang2025colorbench, butt2025gencolorbench}.
Capturing the nuanced association between colors and concepts is essential for achieving semantic alignment with human perception.

\begin{wrapfigure}{r}{0.48\linewidth}
    \centering
    \includegraphics[width=\linewidth]{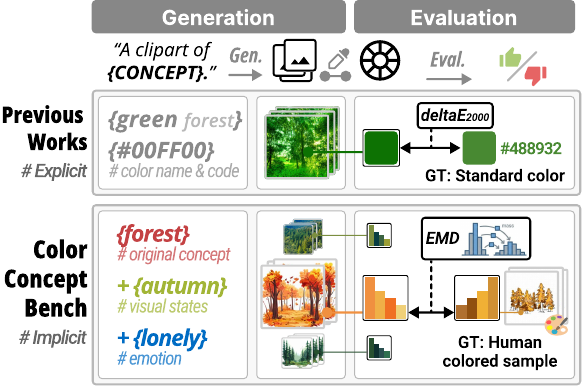}
    \caption{Unlike explicit color matching (top), \bench evaluates implicit semantic alignment using probabilistic color distributions (bottom).}
    \label{fig:colorconceptbench}
    \vspace{-3mm}
\end{wrapfigure}

To evaluate how well T2I models understand color-concept associations, existing approaches typically adopt a two-stage \textit{generate-and-evaluate} pipeline, as illustrated in Figure~\ref{fig:colorconceptbench}.
These methods \citep{lin2024designprobe, tsai2025color, butt2025gencolorbench} rely on explicit color specifications, such as color names (e.g., \q{green}) or color codes (e.g., \q{\#00FF00}) within the text prompt (e.g., \q{A clipart of \{color\} forest}) during the generation phase.
The subsequent evaluation relies on deterministic verification by measuring the distance between the rendered colors and a single, pre-defined ground truth, treating color alignment as a pass/fail check against a reference value. 

However, color is more than a physical attribute, but also a carrier of semantic information in human cognition.
In creative practice, users rarely specify precise color codes; instead, they rely on descriptions of visual states (e.g., \q{autumn}) or emotional atmospheres (e.g., \q{lonely}) to guide generation in prompts \citep{hou2025gencolor}.
Existing benchmarks lack this semantic depth, failing to assess the advanced semantic alignment required for interpreting such implicit visual concepts.
Moreover, humans develop color associations through perceptual experience, which manifests as a \textit{probabilistic distribution of color expectations} rather than a fixed, one-to-one mapping \citep{schloss2024color}.
In contrast, existing approaches reduce the nuanced semantic landscape to a single point, discarding essential information about both diversity and associative intensity.

To bridge these gaps, we introduce \emph{ColorConceptBench}, a new benchmark for evaluating the probabilistic alignment of implicit color semantics in T2I models.
We construct a human-grounded dataset comprising 1,281 concepts and 6,584 human-annotated samples, collected from a controlled sketch colorization task performed by 151 designers to ensure collective consensus.
Accordingly, \bench provides a comprehensive evaluation framework that utilizes distribution-based metrics (e.g., EMD) to quantify the alignment between model-generated color profiles and human ground truth.
\begin{table}[t]
\centering
\caption{Comparison of existing Text-to-Image (T2I) benchmarks.}
\label{tab:benchmark_comparison}
\scriptsize
\setlength{\tabcolsep}{1.2pt}  
\renewcommand{\arraystretch}{1.2}
\begin{tabular}{lllll}
\toprule
\textbf{Benchmark} & \textbf{Scale} & \textbf{Data Format} & \textbf{Data Collection}& \textbf{Focus} \\
\midrule
GenEval \citep{ghosh2023geneval}              & 553         & Text prompts                     & Manual curation & Compositionality\\
T2I-CompBench++ \citep{huang2025t2i}          & 8,000       & Text prompts                     & Manual curation  &Compositionality\\
DPG-Bench \citep{hu2024ella}                  & 1,065       & Text prompts                     & Manual curation &Prompt Adherence\\
Commonsense-T2I \citep{fu2024commonsense}     & $\sim$1,000 & Adversarial prompt pairs         & Expert annotation &Commonsense Reasoning\\
WISE \citep{niu2025wise}                      & 1,000       & Text prompts                     & Manual curation &Commonsense Reasoning\\
MMMG \citep{luo2025mmmg}                      & 4,456       & Image-prompt pairs + KGs         & Expert annotation &Disciplinary Knowledge\\
OneIG-Bench \citep{chang2025oneig}            & 2,440       & Text prompts                     & Real-world + Manual & Compositionality\\
DrawBench \citep{saharia2022photorealistic}   & 200         & Text prompts                     & Manual curation & Compositionality\\
EvalMuse \citep{han2024evalmuse}              & 40K         & Image-text pairs + scores        & Human annotation & Compositionality\\
ColorBench \citep{liang2025colorbench}        & 5,814       & Image-text QA pairs              & Programmatic &Color Understanding\\
GenColorBench \citep{butt2025gencolorbench}   & 44,464      & Text prompts                     & Systematic curation &Color Understanding\\
\textbf{ColorConceptBench (Ours)}             & 6,584       & Colored sketches                         & Expert annotation & Color Understanding\\
\bottomrule
\end{tabular}
\vspace{-1.5em}
\end{table}

We conduct extensive evaluations of nine leading T2I models.
Our analysis reveals a shortcoming of model insensitivity to abstract concepts: while models can reproduce object colors, they consistently struggle to infer appropriate colors from abstract variants (e.g., visual states and emotions).
Furthermore, we show that simply increasing guidance strength does not resolve this gap. 
The findings underscore implicit color alignment as a persistent challenge.

Our contributions are summarized as follows:
\begin{itemize}
    \item \textbf{Human-Grounded Color-Concept Association Benchmark:} 
    We introduce a new benchmark grounded in professional designer annotations, including 6,584 sketch colorizations for 1,281 color concepts, to quantify the probabilistic gap between AI-generated color profiles and human color-concept associations.

    \item \textbf{Probabilistic Evaluation Protocol:} 
          We establish an evaluation protocol that includes both probabilistic and deterministic feature alignment, providing a granular framework to guide future research in improving semantic color controllability.
    
    \item \textbf{Systematic Evaluation and Insights:} 
    We conduct a comprehensive evaluation of leading T2I models across varied concepts, styles, and guidance scales. 
    Our analysis reveals that current models lack sensitivity to implicit semantics, a limitation that remains resistant to stronger guidance.
\end{itemize}
 
\section{Related Work}

\textbf{Color-Concept Association}.
Color is a fundamental semantic channel, conveying emotional tones and cultural associations beyond mere visual appearance~\citep{wierzbicka1990meaning}.
The accurate association of color with concept is critical for applications such as graphic design~\citep{jahanian2017colors} and interior design~\citep{hou2024c2ideas}.
Research supporting these applications is grounded in empirical data collection, where prior work has pursued through two primary approaches.
The first involves human annotation or preference ranking, where participants directly identify or order colors for given concepts~\citep{rathore2019estimating,mukherjee2024llm,volkova2012clex}.
However, this method is often constrained by small scale and labor intensity.
The second approach overcomes this limitation by employing automated or generative methods to extract color–concept associations at scale~\citep{bahng2018coloring,hou2025gencolor}.

While generative model-based approaches show promise, they are largely constrained to colors inherent to object appearance, such as \q{red} for \q{apple} or \q{green} for \q{grass} \citep{setlur2015linguistic}. 
This focus limits their semantic diversity, failing to capture the implicit color associations essential for abstract concepts.
For instance, when prompted with implicit concepts like \q{lonely} or \q{festive}, models often produce inconsistent or semantically misaligned colors, failing to reflect the emotional or cultural palettes that humans intuitively expect; see Appendix Figure~\ref{fig:appendix_modelgen} for example.
To address this limitation, our work introduces a benchmark centered on implicit color concepts, including abstract visual states and emotional associations, enabling a more comprehensive evaluation of high-level semantic alignment in T2I generation.

\noindent
\textbf{Color Evaluation in Generative Models}.
Color is a critical dimension for ensuring both visual realism and semantic alignment in T2I models.
Existing color evaluations primarily focus on attribute binding \citep{butt2024colorpeel,samin2025colorfoil}, assessing whether a model correctly associates a specified color with a target object, using metrics like CLIPScore \citep{hessel2021clipscore} or VQA accuracy \citep{liang2025colorbench}.
A more granular line of work performs fine-grained color verification, testing pixel-level rendering of specific color names or hexadecimal codes \citep{bahng2018coloring,butt2025gencolorbench,tsai2025color}. 
Recent studies have started exploring implicit concepts like emotion and culture in generated images, such as to evaluate the cultural competence \citep{senthilkumar2024beyond} and emotional control \citep{dang2025emoticrafter,yang2024emogen}.

These approaches typically reduce to a deterministic check, using metrics like Euclidean distance in RGB space, to determine if a generated object’s dominant color matches a singular reference value.
This deterministic approach, however, is misaligned with the nature of color semantics that are probabilistic distributions, not deterministic point values \citep{schloss2024color}.
To address this limitation, our approach shifts from singular reference colors to human-grounded color distributions.
We crowdsource representative colorization from multiple designers for each target concept and extract their collective color profiles, capturing the varied ways humans visualize abstract ideas.
This allows us to evaluate alignment using distribution-based metrics, such as Earth Mover’s Distance (EMD) \citep{rubner2000earth}, to measure how closely a model’s generated color distribution matches the human perceptual ground truth.
\begin{figure*}[t]
    \centering
    \includegraphics[width=0.998\linewidth]{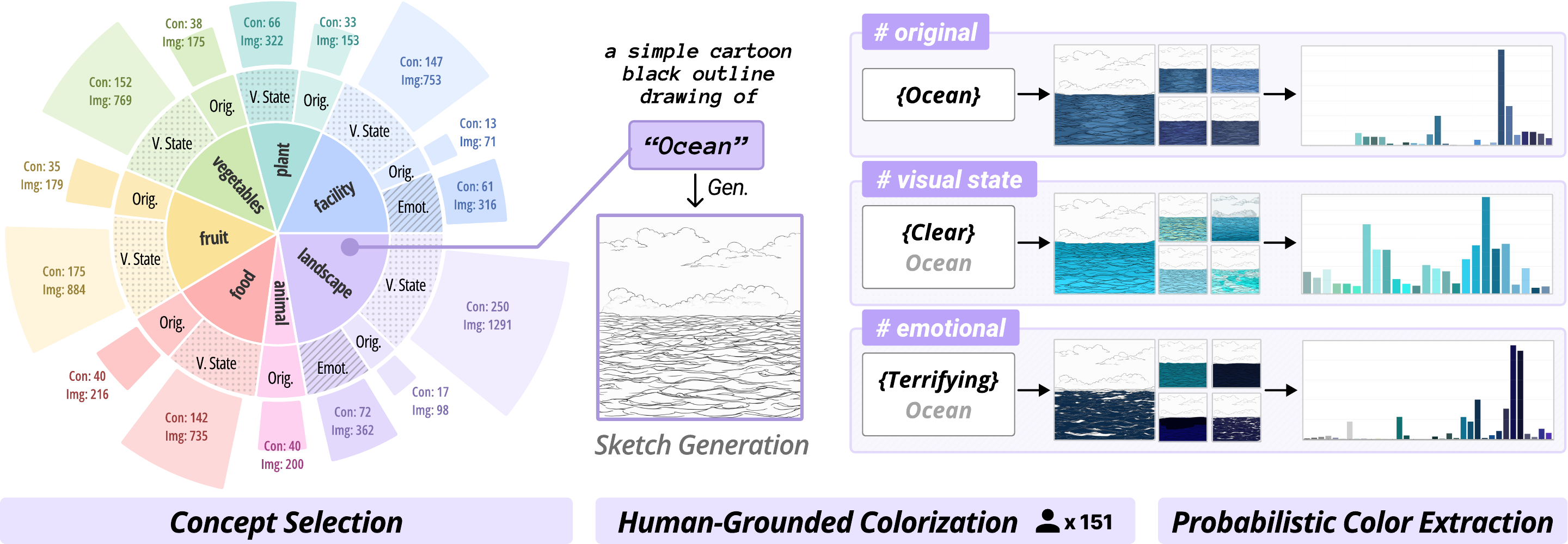}
    \caption{Dataset Statistics and Construction Pipeline. An overview of the hierarchical concept distribution and our three-stage construction process: concept selection, human-grounded colorization, and probabilistic color extraction.}
    \label{fig:dataset-distribution}
\end{figure*}
\section{\bench}
\label{sec:benchmark}

\bench is a benchmark for evaluating probabilistic color-concept understanding in T2I models.
Formally, let \(\mathcal{C}\) denote a set of semantic concepts (e.g., \q{lonely}, \q{festive}).
For each concept \(c \in \mathcal{C}\), we define a \emph{human-grounded color} as a probability distribution over a color space \(\Omega\):
\[
P_{\text{H}}(x \mid c), \quad x \in \Omega,
\]
where \(\Omega\) is a suitable color space.
This distribution is constructed empirically from a curated dataset of human visual annotations of \(c\).
% can be removed if not necessary.
The construction of \bench is illustrated in Figure~\ref{fig:dataset-distribution}.
Our code and dataset are publicly available at \href{https://huggingface.co/datasets/ColorConceptBench/ColorConceptBench}{ColorConceptBench/Dataset\&Code}.
% goes through concepct selection (Sect.~\ref{ssec:concept_selection}), human-grounded colorization (Sect.~\ref{ssec:manual_colorization}), and probalistic color extraction (Sect.~\ref{ssec:color_extract}).

\subsection{Concept Selection}
\label{ssec:concept_selection}
\vspace{-2mm}
Color requires a visual carrier in a generated image.
As such, we begin by selecting a set of target concepts that correspond to concrete objects, which serve as these carriers.
We source these object concepts from the THINGS dataset \citep{hebart2019things}, a large and systematically curated collection of visually grounded entities.
To ensure linguistic relevance and common usage, we apply a subsequent filter based on word frequency from the COCA\footnote{https://www.english-corpora.org/coca/} to ensure relevance and coverage.
% This two-stage process yields a set of 216 concepts covering 6 categories.
% 6 categories? Figure 2 shows 7+1 blank categories.

To evaluate models’ understanding of implicit color concepts in practical contexts, we further define a set of descriptive adjectives that modify entity-based concepts.
Based on the characteristics of each entity category, we first source relevant adjectives from lexical databases and combine them with the original concepts.
Detail links are provided in Table~\ref{tab:lexicon}.
The selected attribute dimensions are divided into two categories:

\begin{itemize}
\item 
\textbf{\textit{Visual State}} describes attributes perceived through observation, such as \q{polluted} or \q{clear} water, and \q{unripe} plants.

\item
\textbf{\textit{Emotional}} covers less-explored affective dimensions, including moods like \q{cozy} or \q{lonely}.
We restrict \emph{emotional} assignments to \emph{facility} and \emph{landscape} entities, as these contexts naturally support affective interpretation.

\end{itemize}

Then, all the adjectives are collected and refined through iterative review with collaborating designers to ensure both linguistic and visual validity.
After combining the original objects with these adjectives, we obtained a final set of 1,281 unique concepts distributed across 7 distinct categories.
The statistics of concepts in each object type and attribute dimensions are listed in Figure~\ref{fig:dataset-distribution}, with the list of concepts provided in Appendix~\ref{sec:concept_taxonomy}.

%%%%%%%%%%%%%%%%%%%%%%%%%%%%%%%%%%%%%%%%%%%%%
%%%%%%%%%%%%%%%%%%%%%%%%%%%%%%%%%%%%%%%%%%%%%
\subsection{Human-Grounded Colorization}
\label{ssec:manual_colorization}
% After defining the set of concepts, we construct a human-annotated color–concept association ground truth dataset.

\vspace{-2mm}
\noindent
We use human coloring as the primary source of ground truth to capture concept-level color associations under controlled conditions. 
This approach helps mitigate uneven concept coverage and noise introduced by lighting, photographic style, and background context in Internet-derived color distributions. We also conduct a preliminary study to explore the limitations and potential of Internet images as a source of ground truth for this task (see Appendix~\ref{sec:appendix_internet_derived}).

\textbf{Sketch Generation}.
After defining the concept set, we generate simple sketches (rather than real images) for each concept, to minimize stylistic variability, reduce background noise, and isolate color as the primary variable of interest.
Using Qwen-Image \citep{wu2025qwen} and Stable Diffusion 3.5 Medium \citep{esser2024scaling}, we generate five sketches per concept with each model. 
A collaborating artist then manually selects the cleanest and most unambiguous sketch for each concept to \mbox{ensure clarity in subsequent color annotation.}

\vspace{1mm}
\noindent
\textbf{Human Colorization}.
Next, we invite professional designers to colorize the selected sketches.
Each concept is independently colored by at least five designers, who are instructed to base their color choices on real-world, intuitive associations drawn from everyday experience with the concept, rather than on personal artistic style.
We recruit designers for their acute color sensitivity, which reduces annotation noise and yields more reliable human ground truth.
The full instructions and designer demographics are documented in Appendix~\ref{sec:human_annotation}.
Finally, we review all submissions and remove clear outliers to ensure that the collected color data reflects stable, semantically grounded human associations.
The quality control protocol and validation results are detailed in Appendix~\ref{sec:quality_control}.
In total, we recruit 151 designers for colorization, each coloring 40 to 50 sketches.
Following our quality control protocol, we retain 6,584 high-quality colored results for the final dataset (see Appendix Figure~\ref{fig:appendix_dataset} for example).

%%%%%%%%%%%%%%%%%%%%%%%%%%%%%%%%%%%%%%%%%%%%%
%%%%%%%%%%%%%%%%%%%%%%%%%%%%%%%%%%%%%%%%%%%%%
\subsection{Probabilistic Color Extraction}
\vspace{-2mm}
\label{ssec:color_extract}
\paragraph{Segmentation.}
We first use Grounding DINO \citep{liu2024grounding} to identify the bounding box of the target concept. The detected bounding box is then passed to the Segment Anything Model (SAM) \citep{kirillov2023segment}, which generates a binary mask of the target concept (See Figure~\ref{fig:appendix_segpipeline}). 
This step ensures that the subsequent analysis is derived strictly from pixels belonging to the concept.
To ensure the reliability of the extracted color information, we evaluate the success rate of the segmentation pipeline and filter out failure cases where the mask captures the background while omitting the target concept. 
Further details on the filtering protocol and success rate statistics are provided in Appendix~\ref{para:seg_filter}.

\paragraph{Quantization and Refinement.}
To capture the primary visual impression, we extract representative colors in the CIELAB space.
However, since a single concept often exhibits multiple color distributions (e.g., \q{apple} can be distinctively red or green), simply averaging all samples would result in inaccurate representations.
To address this, we implement an adaptive grouping strategy:
\begin{enumerate}
\item 
\textit{Perceptual Grouping.} We first cluster images into distinct visual groups based on the CIE $\Delta$E2000 distance between their dominant colors, ensuring that visually disparate modes are processed separately.

\item 
\textit{Refined Merging.} Within each group, we perform a merging process where color centers indistinguishable to the human eye are combined.
The final distribution $P_{H}(x \mid c)$ over the color space $\Omega$ is constructed by aggregating these refined color distributions, weighted by the population of each visual group. 

\end{enumerate}

Implementation details are provided in Appendix~\ref{sec:color_extraction}.

\section{Experiment}

% To evaluate different T2I models, 
% we generate a set of images for each model and estimate the corresponding \emph{model-predicted color distribution} $P_M(x \mid c)$ for each concept $c$.
% Together with the human-grounded color distribution $P_H(x \mid c)$, we quantify the alignment between model outputs and human perceptual expectations.
\vspace{-2mm}
\subsection{Models}
\vspace{-2mm}
We evaluate the performance of nine well-known open-source T2I models on the benchmark.
These models include Stable Diffusion (SD) XL \citep{podell2023sdxl}, SD 3 and 3.5 \citep{esser2024scaling} from Stability AI, Flux.1-dev \citep{flux2024}, Qwen-Image \citep{wu2025qwen}, OmniGen \citep{xiao2024omnigen}, OmniGen2 \citep{wu2025omnigen2}, PixArt-$\alpha$ \citep{chen2023pixartalpha}, and SANA-1.5 \citep{xie2024sana}.
\vspace{-3mm}
\subsection{Implementation}
\vspace{-2mm}
For each concept \(c \in \mathcal{C}\), we construct a diverse image set by varying inference parameters to ensure robustness.
To determine if color associations are style-dependent, we generate images across two styles (\textit{natural} and \textit{clipart}) using standardized templates.
Furthermore, to analyze the impact of generation constraints, we sample images across 7 distinct Classifier-Free Guidance (CFG) scales.
For each combination concept, style, and guidance scale, we generate 5 independent samples at $1024 \times 1024$ resolution.
The prompt templates and detailed generation configurations are provided in Appendix~\ref{sec:appendix_model_inference} and~\ref {sec:appendix_prompt_templates}.
Next, to derive the probability distribution $P_{\text{M}}(x \mid c)$ from these generated images, we employ the identical probabilistic color extraction pipeline described in Sect.~\ref{ssec:color_extract}
\vspace{-3mm}
\subsection{Metric}
\vspace{-2mm}
% \subsubsection{Probabilistic Distribution Alignment}
We employ the probabilistic distribution-based metrics to quantify the statistical and perceptual alignment between these distributions.
For computational convenience, let $p \in \mathbb{R}^K$ and $q \in \mathbb{R}^K$ denote the discrete probability vectors corresponding to $P_{\text{H}}$ and $P_{\text{M}}$ respectively, over the $K$ bins of the UW71 color space \citep{hu2022self, rathore2019estimating}.

\paragraph{Pearson Correlation Coefficient (PCC).}
PCC measures the linear correlation between the color probability distribution perceived by human and that learned by the model. 
% It assesses whether the model preserves the relative proportions of color association.

\begin{equation}
\small
PCC(p, q) = \sum_{k=1}^{K}\frac{ (p_k - \bar{p})(q_k - \bar{q})}{\sqrt{ \sum_{k=1}^{K}(p_k - \bar{p})^2} \sqrt{ \sum_{k=1}^{K}(q_k - \bar{q})^2}},
\end{equation}

where $\bar{p}$ and $\bar{q}$ denote the mean probabilities of the respective distributions.

\paragraph{Earth Mover's Distance (EMD).}
EMD accounts for the perceptual distance between color distributions.
It computes the minimum cost required to transform $p$ into $q$, using a ground distance matrix $D$, where $d_{k_1k_2}$ represents the CIELAB Euclidean distance between color bins $k_1$ and $k_2$:

\begin{equation}
\small
EMD(p, q) = \min_{f} \sum_{k_1=1}^{K} \sum_{k_2=1}^{K} f_{k_1k_2} d_{k_1k_2},
\end{equation}

subject to flow constraints $\sum_{k_2} f_{k_1k_2} = p_{k_1}$, $\sum_{k_1} f_{k_1k_2} = q_{k_2}$, and $f_{k_1k_2} \geq 0$. 
% This metric is particularly robust for capturing perceptual similarity across color shifts.

\paragraph{Entropy Difference (ED).}
To evaluate whether a generative model captures the complexity and diversity of human color concept associations, we compute the absolute difference in Shannon entropy between the two distributions, as: 

\begin{equation} 
\small
ED(p, q) = |\sum_{k=1}^{K} p_k \log(p_k)-\sum_{k=1}^{K} q_k \log(q_k )|. 
\end{equation}

% \subsubsection{Deterministic Feature Alignment}
% While distribution metrics capture the global similarity, they may not explicitly reflect whether the most representative colors are accurate. 

\begin{table}
\scriptsize
\centering 
\caption{Comparison of \textit{probabilistic distribution alignment} of different T2I models. The best and second best results in each column are marked in \textbf{bold} and \underline{underlined}, respectively.}
\setlength{\tabcolsep}{1.2pt}  
\renewcommand{\arraystretch}{1.2}  
\begin{tabular}{c ccc ccc ccc ccc ccc ccc}    
\toprule    
\multirow{3}{*}{\textbf{Method}}&\multicolumn{6}{c}{\textbf{Original}}&\multicolumn{6}{c}{\textbf{Visual state}}&\multicolumn{6}{c}{\textbf{Emotional}}\\    &\multicolumn{3}{c}{\textbf{Natural}}&\multicolumn{3}{c}{\textbf{Clipart}}    &\multicolumn{3}{c}{\textbf{Natural}}&\multicolumn{3}{c}{\textbf{Clipart}}    &\multicolumn{3}{c}{\textbf{Natural}}&\multicolumn{3}{c}{\textbf{Clipart}}\\    \cmidrule(lr){2-4}\cmidrule(lr){5-7}\cmidrule(lr){8-10}\cmidrule(lr){11-13}\cmidrule(lr){14-16}\cmidrule(lr){17-19}    & PCC$\uparrow$& EMD$\downarrow$ & ED$\downarrow$& PCC$\uparrow$& EMD$\downarrow$ & ED$\downarrow$ & PCC$\uparrow$& EMD$\downarrow$ & ED$\downarrow$& PCC$\uparrow$& EMD$\downarrow$ & ED$\downarrow$& PCC$\uparrow$& EMD$\downarrow$ & ED$\downarrow$& PCC$\uparrow$& EMD$\downarrow$ & ED$\downarrow$ \\    \midrule     
Flux.1-dev&\textbf{0.369}&\underline{29.165}&0.621&\underline{0.391}&\underline{26.915}&\underline{0.578}&\underline{0.378}&28.413&0.596&0.337&28.944&\underline{0.501}&\underline{0.407}&28.718&0.771&0.344&27.959&\textbf{0.451}\\     
OmniGen&0.348&30.866&\underline{0.575}&0.363&31.413&0.581&0.319&32.658&0.974&0.278&33.840&0.761&0.342&32.201&1.258&0.290&33.319&0.925\\     
OmniGen2&0.355&30.066&0.648&0.321&36.349&0.845&0.320&31.572&0.960&0.224&39.130&1.463&0.348&30.346&1.090&0.261&35.260&1.447\\     
PixArt-$\alpha$&0.351&30.867&0.596&0.360&29.663&0.607&0.360&29.747&0.577&0.333&30.554&0.536&0.389&30.032&0.763&\underline{0.425}&\underline{26.070}&0.528\\     
Qwen-Image&0.316&30.311&0.696&0.326&30.577&0.774&0.319&30.600&0.710&0.280&31.676&0.633&0.381&28.129&0.660&0.287&31.981&0.501\\     
SD 3&0.348&\textbf{28.839}&0.622&0.359&29.707&0.657&0.338&28.876&0.583&0.276&32.192&0.642&0.374&\textbf{27.379}&\underline{0.621}&0.297&30.748&0.572\\     
SD 3.5&0.321&29.442&0.615&0.325&30.206&0.693&0.308&30.383&\underline{0.545}&0.287&31.050&0.616&0.329&29.108&0.636&0.291&31.768&0.605\\     
SD XL&0.308&29.273&\textbf{0.546}&0.344&27.191&0.609&0.358&\textbf{26.632}&\textbf{0.481}&\underline{0.353}&\textbf{26.299}&0.513&0.378&\underline{27.415}&\textbf{0.477}&0.368&27.412&0.540\\     
Sana-1.5&\underline{0.360}&29.425&0.596&\textbf{0.428}&\textbf{26.151}&\textbf{0.524}&\textbf{0.390}&\underline{27.861}&0.552&\textbf{0.375}&\underline{28.243}&\textbf{0.442}&\textbf{0.425}&28.033&0.694&\textbf{0.438}&\textbf{25.058}&\underline{0.457}\\    
\bottomrule  
\end{tabular}  
\label{tab:distribution_metric}
\end{table}

\begin{figure}[b]
  \centering
  \includegraphics[width=0.8\textwidth]{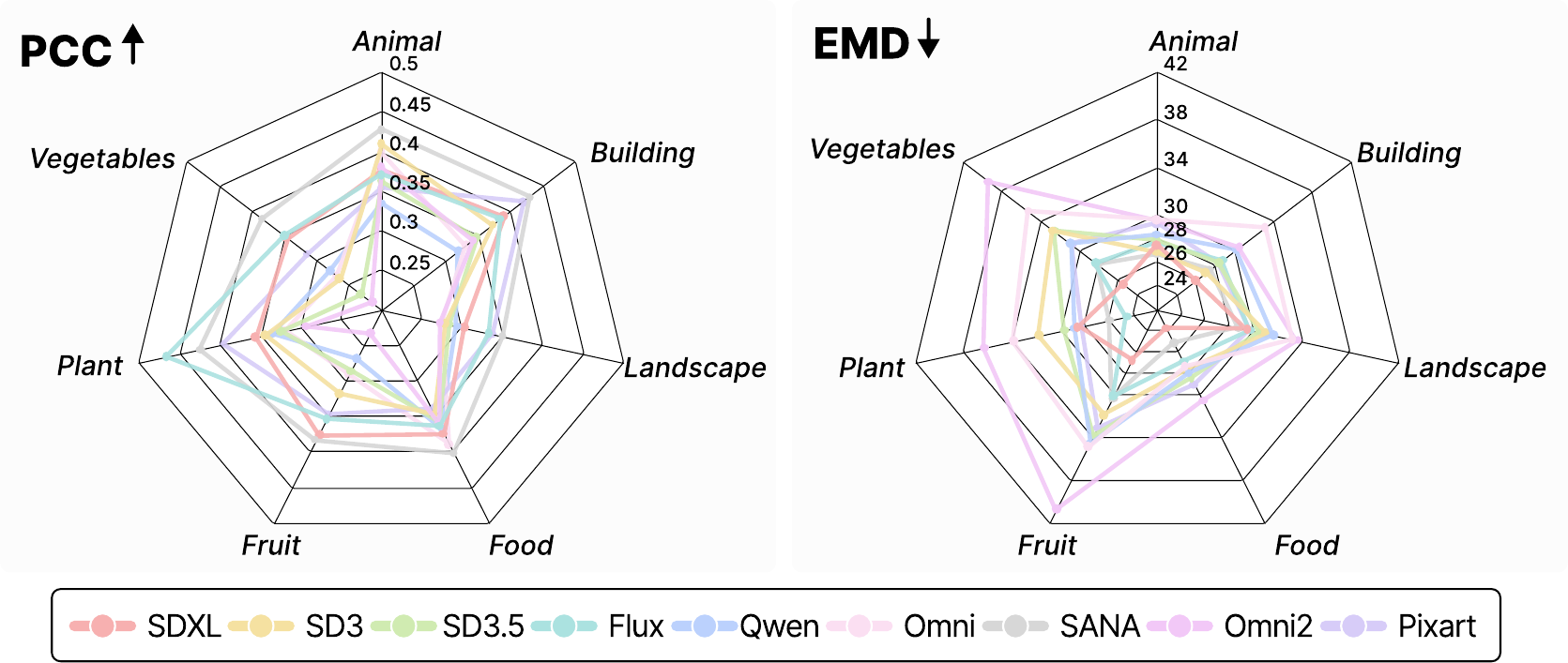}
  \caption{Comparison of probabilistic distribution alignment of different T2I models across categories.}
  \label{fig:result_cate}
  % \vspace{-3mm}
\end{figure}

\begin{figure}[htb]
  \centering
  \includegraphics[width=0.988\linewidth]{fig/cases-ring_compressed.pdf}
  \caption{Qualitative comparison of color-concept association across different text-to-image models. Colors shift for base nouns (e.g., `cabin') and modified concepts involving visual states (e.g., `autumnal') or emotions (e.g., `lonely'), across both natural and clipart styles, shown with color distribution and dominant colors with sample number.}
  \label{fig:results_cases}
  \vspace{-2mm}
\end{figure}
\vspace{-1mm}
\subsection{Results}
\vspace{-2mm}
\textbf{Overall Performance.}
Table~\ref{tab:distribution_metric} presents the probabilistic distribution alignment of T2I models across concept categories and visual styles.
Overall, Sana-1.5, Flux.1-dev, and Stable Diffusion XL consistently rank among the top performers across most conditions. 
% Notably, for a given model, alignment is higher for Original concepts than for their variants Visual State and Emotional, and higher for Clipart than for Natural imagery.
However, the effects of modifiers depend on both modifier type and visual style.
Comparing the same base concepts with and without modifiers, we find that Emotional modifiers generally improve color alignment in Natural images, while their effects in Clipart images are mixed.
Detailed results are provided in Tables~\ref{tab_matched_visual} and~\ref{tab_matched_emotional}.

Surprisingly, Stable Diffusion XL demonstrates exceptional performance on Visual State concepts, achieving the best EMD scores in both Natural and Clipart styles.
Meanwhile, Stable Diffusion 3 stands out on Natural imagery, achieving the best EMD scores for both Original and Emotional concepts in this style.
Among all models, Sana-1.5 (4.8B) achieves state-of-the-art performance. 
This indicates that Sana-1.5, despite its smaller parameter count compared to larger models like Qwen-Image (20B), excels at mapping textual semantics to corresponding color distributions across all concepts.

\textbf{Category-level analysis.}
Figure~\ref{fig:result_cate} presents the per-category breakdown of PCC and EMD scores across all models.
Overall, model performance varies substantially across base categories, revealing how strongly object-color priors in training data influence generative color alignment.
Categories with strong intrinsic color priors yield higher alignment.
Models consistently achieve higher PCC and lower EMD on categories such as \textit{Animal} and \textit{Plant}, where real-world color distributions tend to be semantically grounded and relatively predictable (e.g., green bush).
The relative uniformity of these color distributions makes it easier for models to retrieve and reproduce semantically appropriate colors.
In contrast, \textit{Landscape} yields the lowest PCC scores across nearly all models, likely because landscape imagery encompasses highly diverse color palettes depending on time of day, season, and geographic context, making it difficult for models to converge on a consistent color expectation.
Despite this category-level variation, the relative ranking of models remains largely stable across all categories.
Sana achieves the best overall alignment under both PCC and EMD, followed by Flux and Stable Diffusion XL, while OmniGen2 consistently underperforms, recording the highest EMD values across nearly all categories and indicating a systematic difficulty in reproducing semantically grounded color distributions regardless of object type.

% \vspace{1mm}
% \noindent
\textbf{Qualitative Results.}
Figure~\ref{fig:results_cases} presents qualitative examples generated by different models across various concepts and styles.
These examples align with the quantitative findings discussed earlier. 
For instance, outputs from OmniGen, which show weak human alignment in the metrics, demonstrate a stronger reliance on form and composition than on color matching, particularly for the Visual State and Emotional concepts.
\vspace{-1em}
\subsection{Human Judgment}
\vspace{-2mm}
To validate that our probabilistic distribution-based metrics better reflect human perceptual judgment than deterministic alternatives, we conducted a perceptual study to measure their consistency with human judgment.
Here, deterministic alternatives refer to metrics that assess performance based on single-color accuracy, such as Dominant Color Accuracy (DCA) and Hue Angular Difference ($\Delta$ Hue); details are provided in Appendix~\ref{sec:appendix_deterministic}.
We randomly sampled 150 concepts and generated outputs using three representative models: Sana, Stable Diffusion XL, and OmniGen.
This experiment employs a pairwise comparison protocol hosted on a Gradio interface \citep{abid2019gradio}.
To ensure statistical robustness, we invited 62 participants, where each pair was evaluated by five independent annotators.
\begin{wraptable}{r}{0.6\linewidth}
\vspace{-3mm}
\centering
\caption{Validation of metric alignment with human judgment. Metrics based on probabilistic distributions, especially EMD, align more closely with human judgment than those based on deterministic features.}
\scriptsize
\begin{tabular}{lcccc}
\toprule
\textbf{Type}&\textbf{Metric} & \textbf{Kendall's Tau} & \textbf{Spearman} & \textbf{Agreement} \\
\midrule
\multirow{3}{*}{Probabilistic}&PCC & 0.230 & 0.316  & 63.65\% \\
&EMD & \textbf{0.276} & \textbf{0.378}  & \textbf{68.37\%} \\
&ED & 0.173 & 0.235  & 47.24\% \\
\cmidrule(lr){2-5}
\multirow{2}{*}{Deterministic}&DCA &0.118 & 0.136 & 11.94\% \\
&$\Delta$Hue & 0.162 & 0.219  & 43.44\% \\
\bottomrule
\end{tabular}
\label{tab:correlation}
\end{wraptable}
For each trial, participants are presented with the human-annotated ground truth alongside the images generated by two randomly paired models.
Participants are asked to determine which model's output aligned more closely with the human ground truth.

We quantify alignment between our metrics and human preference using Kendall's Tau ($\tau$) and Spearman ($\rho$) correlation coefficients, and agreement ratio.
Table~\ref{tab:correlation} presents the results, which reveal that the adopted distribution-based metric, especially EMD, exhibits a significantly higher correlation with human judgment compared to deterministic metrics commonly employed in prior studies, which typically assess performance based on single-color accuracy.
Detailed information about human judgment can be found in Appendix~\ref{sec:appendix_metrics}.

%%%%%%%%%%%%%%%%%%%%%%%%%%%%%%%%%%%%%%%%%%%%%%
%%%%%%%%%%%%%%%%%%%%%%%%%%%%%%%%%%%%%%%%%%%%%%
\vspace{-2mm}
\section{Discussion}

\vspace{-2mm}
\subsection{Abstract Concepts are Difficult to Color} 
\vspace{-2mm}
% \textbf{The Concrete-Abstract Performance Gap.}
% A trend observed across most models is the performance degradation from concrete to abstract concepts, namely the \textit{Visual State} and \textit{Emotional} semantic modifiers.
The preceding analysis shows that the effects of semantic modifiers on color alignment vary across models and visual styles. These results raise a further question: to what extent do models adjust the colors associated with a base concept in response to its modifiers?
\begin{wrapfigure}{r}{0.58\linewidth}
    \centering
    \includegraphics[width=\linewidth]{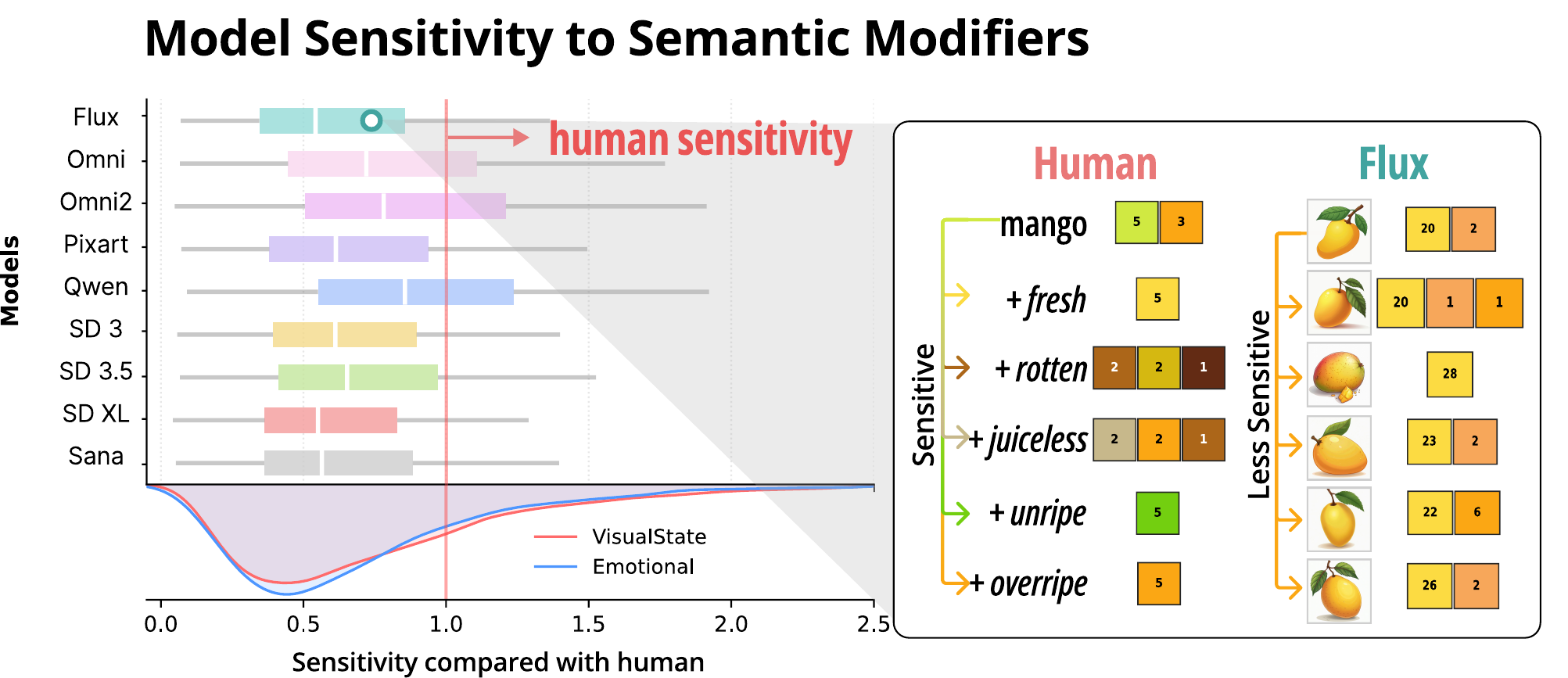}
    \caption{Models consistently exhibit lower color shift magnitudes than the human baseline (left), conservatively prioritizing intrinsic object colors over modifier-induced adjustments (right). Detailed numerical results are provided in Table~\ref{tab_fig5_quartiles}.}
    \label{fig:sensitivity}
\end{wrapfigure}

To address this question, we evaluate the model's semantic sensitivity by measuring the color shift induced by modifiers.
Specifically, we calculate the EMD between the color distribution of a base concept (e.g., `Mango') and its modified version (e.g., `Rotten Mango').
A larger EMD signifies a larger distributional difference, indicating a stronger response to the modifier.
We compare its shift intensity against the human baseline.
For instance, human designers exhibit varying degrees of intensity depending on the modifier (e.g., a drastic palette shift for `rotten' vs. a subtle adjustment for `fresh').
Ideally, models should exhibit a shift intensity comparable to humans; a significantly lower intensity indicates that the model fails to update the color distribution in response to the semantic cue.

However, we find that models are consistently \textbf{under-sensitive}, as shown in Figure~\ref{fig:sensitivity} (left).
Quantitatively, for \textit{Visual State} modifiers (see Appendix Table~\ref{tab:sensitivity_scores_cate}), the average color shift produced by models is only \textbf{$\sim$77\%} of the human shift intensity.
This sensitivity drops further to \textbf{$\sim$72\%} for \textit{Emotional} modifiers.
This indicates a conservative tendency: models prioritize the intrinsic color of the noun (e.g., keeping a mango yellow) rather than making the generative adjustments required by the adjective.

By cross-referencing sensitivity magnitude with visual outputs, we identify three distinct response behaviors:
\textit{(1) Semantic Inertia (Low Sensitivity).} 
Models like Flux fail to override strong object priors, exhibiting negligible response to modifiers.
For instance, distinct prompts like \q{mango} and \q{unripe} mango produce nearly identical color distributions (see Figure~\ref{fig:sensitivity}), indicating a failure to initiate the necessary distributional shift.
\textit{(2) Semantic Over-Correction (High Sensitivity / Drift).} 
Finally, models such as Qwen exhibit excessive sensitivity that can lead to semantic drift (See Appendix Figure~\ref{fig:extra_case}).
While their shift magnitude rivals that of humans, they often lack control. For instance, when generating a ``polluted lake,'' the model may completely overwrite the water's blue tones with mud colors, whereas human annotators typically retain the blue hue while introducing grayish tones. 
Here, high sensitivity reflects a failure to preserve the subject's identity, rather than accurate adaptation and balancing.
\textit{(3) Precise Adaptation (Balanced Sensitivity).} 
In ideal cases, models like SDXL and Sana achieve small but semantically accurate shifts. For the \q{rotten apple} case (Figure~\ref{fig:results_cases}), SDXL effectively shifts the palette from canonical red to brownish-green decay without losing the object's identity. 
Similarly, Sana demonstrates a subtle yet directionally aligned color shift. 
This demonstrates a balance: these models interpret the modifier as a specific physical property update rather than a generic style transfer.

\vspace{-2mm}
\subsection{Inefficacy of Stronger Guidance.}
\vspace{-2mm}
\begin{wrapfigure}{r}{0.58\linewidth}
    \centering
    \includegraphics[width=\linewidth]{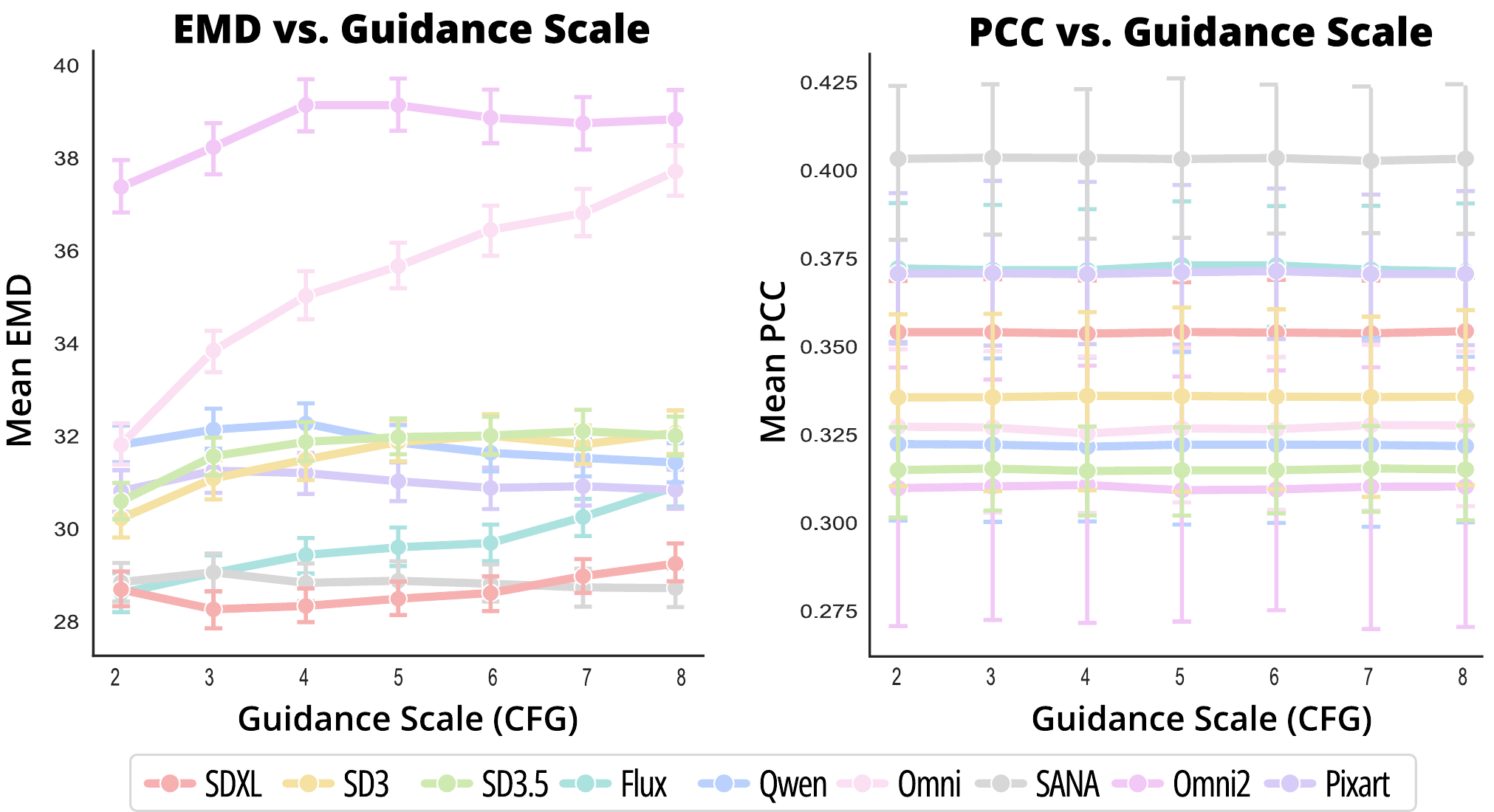}
    \caption{Impact of Guidance Scale. Increasing the CFG scale generally leads to higher EMD across models (worse alignment), while PCC remains largely unchanged. Detailed numerical results are provided in Table~\ref{tab_fig6_emd} and~\ref{tab_fig6_pcc}. }
    \label{fig:cfg_influence}
\end{wrapfigure}

While Classifier-Free Guidance (CFG) is typically increased to enhance alignment with the prompt, our results indicate it generally fails to improve semantic color alignment.
As shown in Figure~\ref{fig:cfg_influence}, the majority of models exhibit no clear benefit from increased guidance strength, with EMD scores varying without a consistent trend across the CFG range of 2 to 8.
For the OmniGen series in particular, performance degrades as guidance increases, suggesting that stronger conditioning actively disrupts the model's color generation rather than refining it.
On the PCC side, scores remain remarkably flat across all models and all guidance levels, indicating that the overall correlation between generated and target color distributions is largely unaffected by inference-time tuning.
This stability suggests that models retain a fixed set of dominant color associations regardless of guidance strength, neither introducing new hues nor suppressing existing ones in response to stronger conditioning.
Together, these findings suggest that implicit color binding behaves as a fixed intrinsic capability that is established during training and cannot be meaningfully improved through stronger CFG at inference time, unlike spatial or compositional adherence, which is known to be responsive to guidance scaling.

\subsection{Towards Better Color–Concept Alignment.}
Larger and more recent models do not consistently achieve better color–concept alignment. Sana-1.5 remains competitive with a compact 4.8B backbone, while Stable Diffusion XL remains among the strongest models under the designer ground truth and closest to the Internet-derived reference distributions. Rankings across architectures are also mixed. These results motivate looking beyond model size, recency, and backbone design.

Two aspects of training offer possible explanations.
Caption quality and text conditioning may help models infer implicit appearance semantics from short prompts, as illustrated by SANA's multi-model captioning and Complex Human Instruction \citep{xie2024sana}.
This is especially relevant to \bench, where prompts never explicitly specify the expected colors.
Besides, post-training objectives often emphasize prompt adherence, composition, editing consistency, and text rendering, which may improve without strengthening typical color associations \citep{wu2025qwen}.
Stable Diffusion XL's performance may reflect preserved pretraining color statistics, while Nano Banana 2's may reflect broader visual world knowledge that includes concept appearance.
Additional results for recent and closed-source models are provided in Appendix~\ref{app:recent_models}.

These explanations suggest two directions for improving color–concept alignment.
\begin{itemize}
\item Appearance-aware captions and prompts. During training, image captions can explicitly state what concepts typically look like. At inference time, CHI-style prompt expansion can add appearance details that are implicit in a short prompt. Both approaches give the model more information about concept appearance without changing the training objective.
\item Color-aware training objectives. Training can reward alignment with human color expectations through signals that directly score the generated colors, such as localized objectives on the target concept's colors and distribution-level objectives that compare multiple generations with human color distributions.
\end{itemize}

\vspace{-3mm}
\section{Conclusion}
\vspace{-1mm}
In this work, we introduce \bench, a benchmark to evaluate color-concept association, an important yet under-explored aspect of text-to-image generation.
After evaluating nine T2I models on 1,281 concepts with 6,584 human annotations, our study reveals that current T2I models still struggle to associate concepts with human-expected colors.
Specifically, we find that models fail to maintain color-concept associations ability as semantic complexity increases.
Moving forward, we plan to extend our investigation to cross-cultural contexts, exploring how human color-concept associations vary across different backgrounds and whether T2I models can capture these culturally specific semantic nuances.
We envision this benchmark as a foundation to foster future advancements in semantic-aware color generation.
\vspace{-1mm}
\paragraph{Limitations.}
There are several limitations of this work.
1) \emph{Demographic and Cultural Scope.}
Our human annotation data is predominantly collected from a single cultural region. 
Since color symbolism and emotional associations are culturally dependent, our current findings represent a specific geo-cultural distribution. 
% Consequently, while this provides a baseline, the observed color-concept associations may not generalize to global contexts with distinct cultural color associations.
2) \emph{Scope of Concept and Semantic Modifiers.}
While our study extensively investigates abstract semantic modifiers, we restrict the base concepts to tangible objects with intrinsic visual properties.
% We cover seven classes of natural and artificial objects, explicitly excluding purely abstract subjects (e.g., \q{happy} or \q{freedom}). 
% This ensures that we measure color shifts relative to a stable structural anchor, thereby avoiding the unbounded visual variability inherent in the generation of abstract concepts.
Regarding the semantic modifier, another limitation is the omission of culturally specific cues.
Color-concept associations are often culture-dependent rather than universal. 
For instance, the concept ``wedding'' is strongly associated with Red in many Eastern cultures (symbolizing luck and joy), whereas it is predominantly linked to White in Western traditions (symbolizing purity).

\section{Acknowledgments}
This work is supported in part by the National Natural Science Foundation of China (62572415) and Guangzhou Basic and Applied Basic Research Foundation (2024A04J6462).

\bibliography{main}
\newpage
\appendix
\section{Appendix: Statements}
\paragraph{Ethics Statement.}
We acknowledge the broader ethical implications of generative AI. 
The study was approved by the authors’ institution’s ethics board, with all participants providing informed consent in data annotation process and human judgment.
Regarding the human evaluation, we ensured that all data was collected anonymously with informed consent obtained from participants.
We confirm that our dataset contains no personally identifiable information (PII) and has been screened to be free of offensive content or societal biases. 
Consequently, the release of this benchmark poses no foreseeable harm to society and does not facilitate the generation of malicious content like deepfakes.
All models used are publicly available.
And the link of our code and dataset has been provided in the paper to ensure reproducibility.
\paragraph{Reproducibility Statement.}
To facilitate reproducibility, we have made the entire dataset, source code, and scripts needed to replicate all results presented in this paper available on Hugging Face. 
Elaborate details of all experiments have been provided in the Appendices.

\paragraph{LLM Usage Statement.}
We used GPT-5.5 solely for grammar correction and language polishing.
The model was not involved in ideation, data analysis, or deriving any of the scientific contributions presented in this work.

% ==========================================
% SECTION A: DATASET CONSTRUCTION
% ==========================================
\section{Dataset Construction Details}
\label{sec:appendix_dataset}

This section provides supplementary details regarding the construction of \bench, including concept selection statistics, the sketch generation pipeline, and the human annotation protocol.
\begin{figure}[htb]
    \centering
    \includegraphics[width=0.38\linewidth]{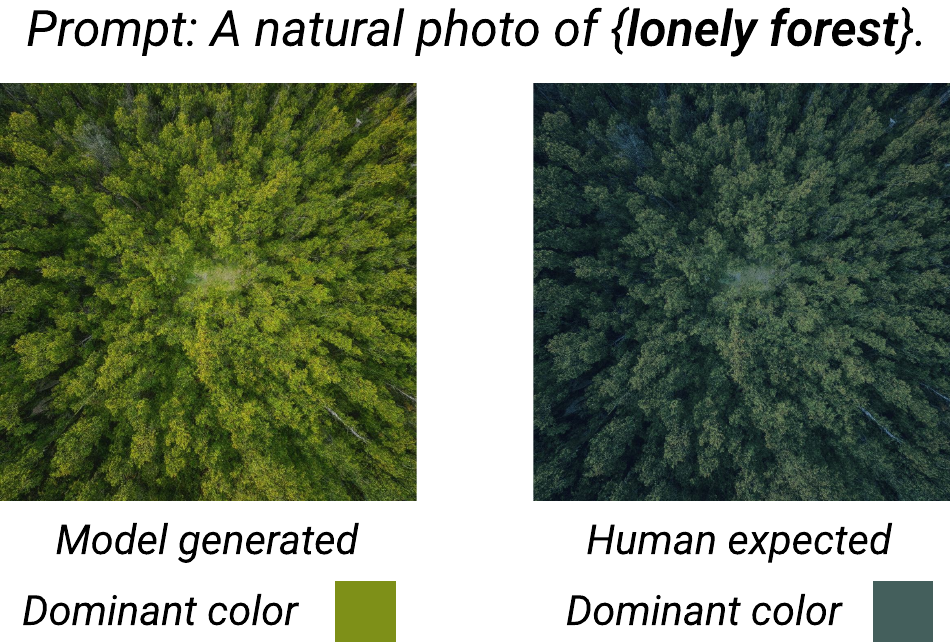}
    \caption{Misalignment color-concept association with human expectation.}
    \label{fig:appendix_modelgen}
\end{figure}
\subsection{Concept Taxonomy \& Statistics}
\label{sec:concept_taxonomy}
The concept categories consist of seven classes, covering both natural and artificial objects:
\begin{itemize}
    \item Vegetables: common vegetables, such as potatoes and bean sprouts.
    \item Fruit: common fruits, such as apple and banana.
    \item Plant: flowers and other plants, such as lily and clover.
    \item Animal: common animals, such as dog and bat.
    \item Food: processed foods and beverages, such as bagel and coffee.
    \item Landscape: natural scenes, such as lake and mountain.
    \item Building: architectural structures, both ancient and modern, such as pyramid and skyscraper.
\end{itemize} 

A concept is defined either as a base word (original concept), a base word with a visual state adjective, or a base word with an emotional adjective.
Visual state adjectives denote changes in the objective physical state of the concept, whereas emotional adjectives capture more abstract qualities, reflecting atmosphere or mood.

Table~\ref{tab:condept_list} provides the detailed concept list.
Each entry corresponds to an original concept and its associated visual state and emotional adjectives.
Concepts are uniquely identified and grouped by category, providing a comprehensive reference for all concepts used in our experiments.

\begin{table}[t]
\scriptsize
\caption{Lexicon for adjectives.}
\centering
\begin{tabularx}{\linewidth}{lX}
\hline
\textbf{Category} & \textbf{Lexicon Link}  \\
\hline
Fruit \& Vegetables &  \url{https://www.words-to-use.com/words/fruits-vegetables/}\\
\cmidrule(lr){2-2}
Water Body & \url{https://www.yourdictionary.com/articles/water-words-descriptive-writing} \\
\cmidrule(lr){2-2}
Weather \& Sky & \url{https://www.writerswrite.co.za/words-to-describe-weather/} \\
\cmidrule(lr){2-2}
Floral &  \url{https://www.slowflowerspodcast.com/wp-content/uploads/Slow-Flowers_Creative_Workshop_Floral-Adjectives_Nouns_Verbs.pdf}\\
\cmidrule(lr){2-2}
Terrain& \url{https://adjectives-for.com/terrain} \\
& \url{https://simplicable.com/descriptive-words/words-to-describe-nature}\\
\cmidrule(lr){2-2}
Food & \url{https://www.words-to-use.com/words/sweets-desserts/}\\
\hline
\end{tabularx}
\label{tab:lexicon}
\end{table}

\begin{figure}[htb]
    \centering
    \includegraphics[width=0.68\linewidth]{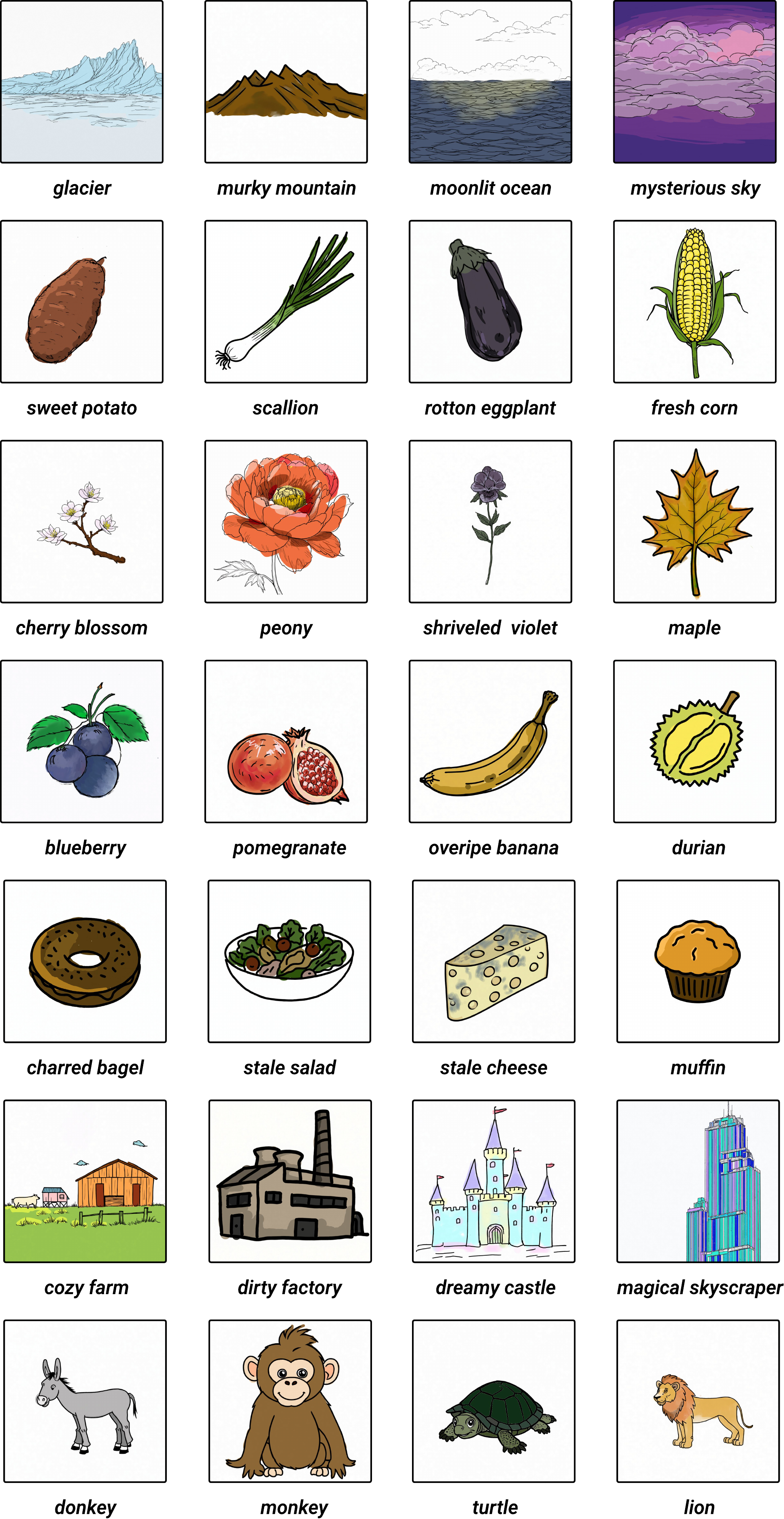}
    \caption{A gallery of our human-annotated dataset.}
    \label{fig:appendix_dataset}
\end{figure}

\subsection{Sketch Generation Pipeline}
% Describe Qwen-Image/SD3.5 usage and filtering here.
\paragraph{Model Selection.}
To generate sketch images that are both visually clear and suitable for color annotation, we employ two complementary text-to-image generation models: Qwen-Image \citep{wu2025qwen} and Stable Diffusion 3.5 Medium \citep{esser2024scaling}. 

Specifically, sketches generated by Qwen-Image tend to exhibit clean and simplified contours with minimal visual clutter, making them highly suitable for preserving object structure and avoiding unintended visual cues. 
However, such sketches may occasionally lack sufficient regions for coloring, limiting their usefulness for certain object categories.
These models provide complementary strengths, allowing us to select sketches that best balance structural clarity and interior detail for color annotation.
\paragraph{Prompt Design.}
All sketches are generated using a unified prompt template designed to suppress color and texture cues while preserving structural information. 
Specifically, we adopt the following prompt formulation:

``A simple cartoon black outline drawing of [original concept], without coloring, without shadow, white background.''

This prompt explicitly enforces the absence of color, shading, and lighting effects, ensuring that the resulting images convey only the structural characteristics of the target concept. 
By standardizing the prompt across all concepts, we minimize stylistic variation that could otherwise influence participants’ color perception.

\begin{figure*}[t]
    \centering
    \includegraphics[width=0.98\linewidth]{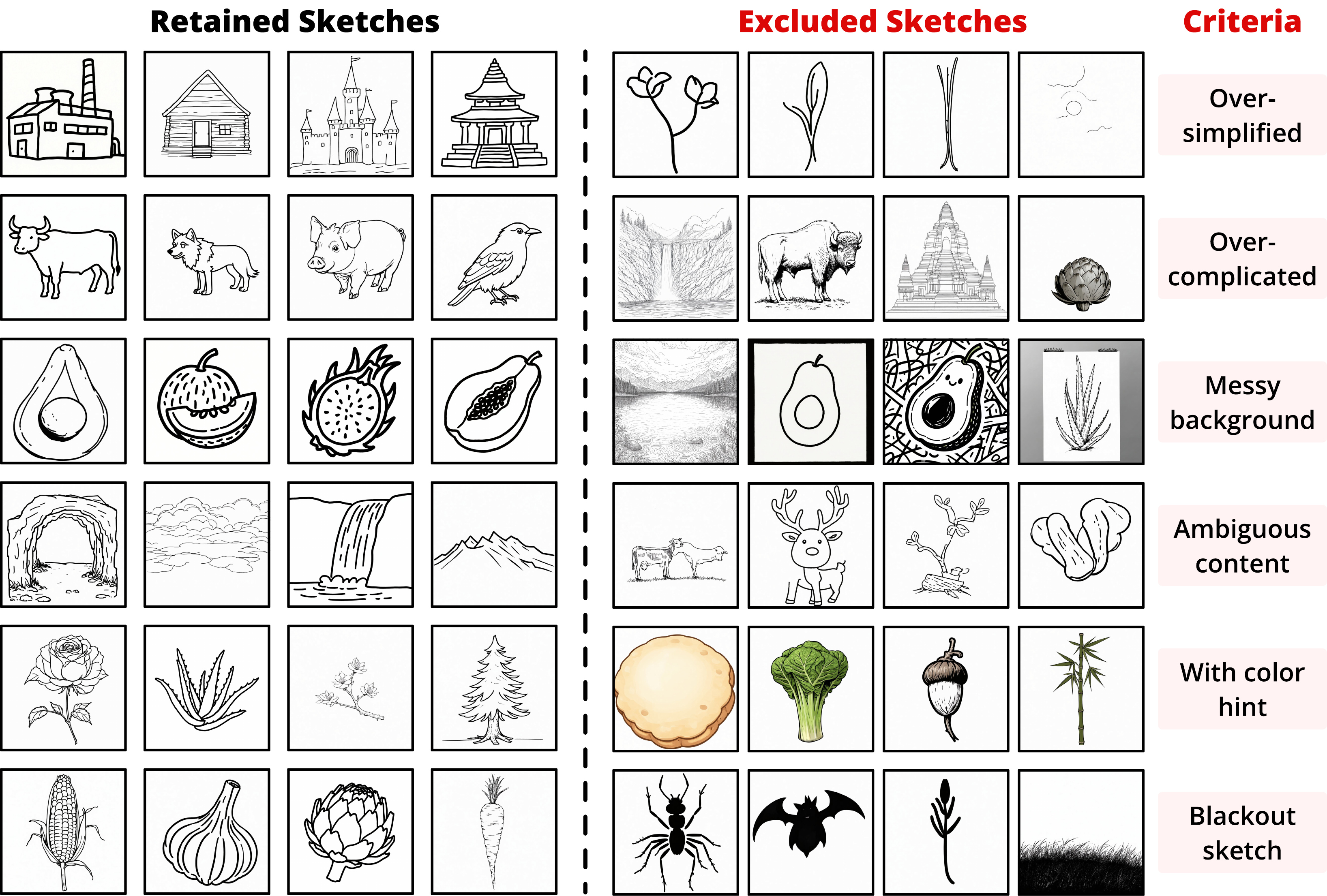}
    \caption{Our sketch}
    \label{fig:appendix_sketch}
\end{figure*}

\paragraph{Generation Settings and Bias Control.}
For the same concept, different visual state or emotional modifiers are applied using the same selected sketch, ensuring consistency across variations.
With 216 original concepts in total, this results in 216 sketches.
However, line drawings can sometimes contain unwanted visual hints due to complex shapes. 
To address this, we manually review and select the sketches. 
For each base concept, we generate five candidates per model using a guidance scale of 5 to 7.
From these, we choose the one with the best structural clarity and no unintended shading or texture.
From the generated candidates, we select one sketch that best satisfies predefined criteria.
Images with artifacts, messy backgrounds, or ambiguous structures are discarded (see Figure~\ref{fig:appendix_sketch} for filtered out examples).
By incorporating controlled multi-sample generation and selection into the pipeline, we reduce the impact of random generation artifacts.
Representative examples of the finalized sketches are provided in Figure~\ref{fig:appendix_sketch}.

\subsection{Human Annotation Process}
\label{sec:human_annotation}
\paragraph{Demographic Information.}
All annotators participating in the study were professionals with a background in design or the visual arts.
Their expertise spans areas such as visual design, industrial design, and fine arts (e.g., painting and illustration).
Therefore, they possess strong color perception skills, ensuring that the annotations are accurate and reliable.

\paragraph{Designer Instructions.} 
% Paste the specific instruction text given to designers here (e.g., "rely on intuitive experience...").
Annotators were provided with detailed instructions to ensure consistent and meaningful color annotations:

\paragraph{1) Color Selection Guidelines.}  
Annotators were asked to use the most common colors associated with each concept according to their own memory and understanding. 
The colors should reflect realistic associations rather than cartoonish or childlike styles.
\paragraph{2) Task-specific color association.}
Colors must be applied strictly according to the given concept. 
Annotators should focus on the concept-relevant parts of the image, leaving unrelated areas uncolored.
For example,for the concept ``grassland'', only the grass should be colored; mountains, sky, and other elements should remain blank.
For the concept ``coffee'', the beverage itself should be colored, but the cup or other containers do not require coloring.
\paragraph{3) Reference examples.}
Example images were provided to help annotators understand the target coloring goals.
\paragraph{4) Consent Form.}
All annotators were required to read and sign an informed consent form before beginning the experiment, ensuring that participation was voluntary and ethically compliant.
\paragraph{5) Pre-experiment practice.}
Before beginning the main annotation task, each annotator completed a small pre-experiment of three images.
These initial annotations were manually reviewed to ensure compliance with the instructions before proceeding to the formal experiment.
\paragraph{Payment.}
We pay each participant \$15 for approximately 30 to 60 minutes of their time and effort.

\subsection{Quality Control.}
\label{sec:quality_control}
% Describe how you filtered outliers (Human Rating Criteria).
To ensure the reliability and semantic validity of our dataset, we implemented a three-stage quality control protocol, combining quantitative consistency checks with expert qualitative verification.
\paragraph{1) Qualitative Review and Iterative Verification.}
We first conducted a manual review process to identify annotations that may deviate from common-sense or domain-consistent interpretations of the corresponding concepts. 
All colorized results were examined by domain experts with experience in visual semantics and design.

For annotations considered potentially implausible or ambiguous, the corresponding participant was asked to recolor the same concept. 
If the two rounds of annotations exhibited consistent color patterns, the result was retained and interpreted as a stable reflection of the participant’s internal conceptual understanding, even when it differed from more frequently observed or prototypical associations.
In such cases, follow-up inquiries were conducted to document the participant’s reasoning and intended interpretation.

If the two rounds showed noticeable discrepancies, we further investigated potential causes such as misunderstanding of the target concept or misinterpretation of the provided prompt.
The participant was then asked to repeat the annotation process until a stable and self-consistent result was obtained. 
This iterative procedure allowed us to filter out accidental or low-quality annotations and ensured that the retained data reliably reflected participants’ genuine conceptual associations.
\paragraph{2) Quantitative Consistency Check.} 
To guarantee high-quality ground truth, each concept was independently annotated by five professional designers who underwent specific training for this task.
Specifically, for each concept $c$, we extracted the color distributions in the UW71 space from the five annotated images, denoted as $\{p_1, \dots, p_5\}$. 
We then computed the average pairwise Earth Mover's Distance (EMD) within the group:
\begin{equation}
\text{EMD}(c) = \frac{1}{10} \sum \text{EMD}(p_i, p_j)
\end{equation}
where the denominator 10 represents the number of unique pairs among the five annotators. 
A lower score indicates higher consensus among the designers.

\begin{table}[t]
%\scriptsize
\caption{Distribution of Expert Agreement Patterns}
\centering
\begin{tabular}{lcc}
\hline
\textbf{Expert Votes} & \textbf{Interpretation} & \textbf{Image Count} \\
\hline
3 Yes / 0 No & Unanimous Consistent & 399 \\
0 Yes / 3 No & Unanimous Inconsistent & 4 \\
2 Yes / 1 No & Majority Consistent & 127 \\
1 Yes / 2 No & Majority Inconsistent & 32 \\
\hline
\textbf{Total} &  & \textbf{562} \\
\midrule
\end{tabular}

\label{tab:expert_agreement}
\end{table}

\paragraph{3) Expert Verification for High-Variance Concepts.} 
Recognizing that abstract or complex concepts may naturally exhibit higher variance (e.g., \textit{`rotten apple'} and \textit{`lonely cabin'}), we perform a targeted review on the ``long-tail'' data. 
We identify the top 10\% of concepts with the highest average EMD scores, indicating the lowest agreement. 
To distinguish between valid semantic ambiguity and potential annotation errors, we recruit a separate panel of trained experts to review these high-variance samples. 
\begin{itemize} 
    \item \textbf{Review Protocol:} For each concept, three independent experts review both the coloring images and color distributions. 
    \item \textbf{Binary Validation:} The experts perform a binary classification task, labeling each copy as either \textit{Consistent} or \textit{Inconsistent} with the semantic meaning of the concept. 
    \item \textbf{Decision Rule:} Annotated images that fail to secure a majority vote are \mbox{removed from the final dataset.}
\end{itemize} 
Finally, we perform quality control on 562 colored images and discard 36 inconsistent samples.
This hybrid approach ensures that our dataset retains rich, diverse color associations for abstract concepts while filtering out low-quality or error annotations.
Samples \mbox{of the verification results are shown in Figure~\ref{fig:qualitycontrol}.}

\begin{figure*}[htb]
\begin{subfigure}{0.45\textwidth}
    \centering
    \includegraphics[width=\linewidth]{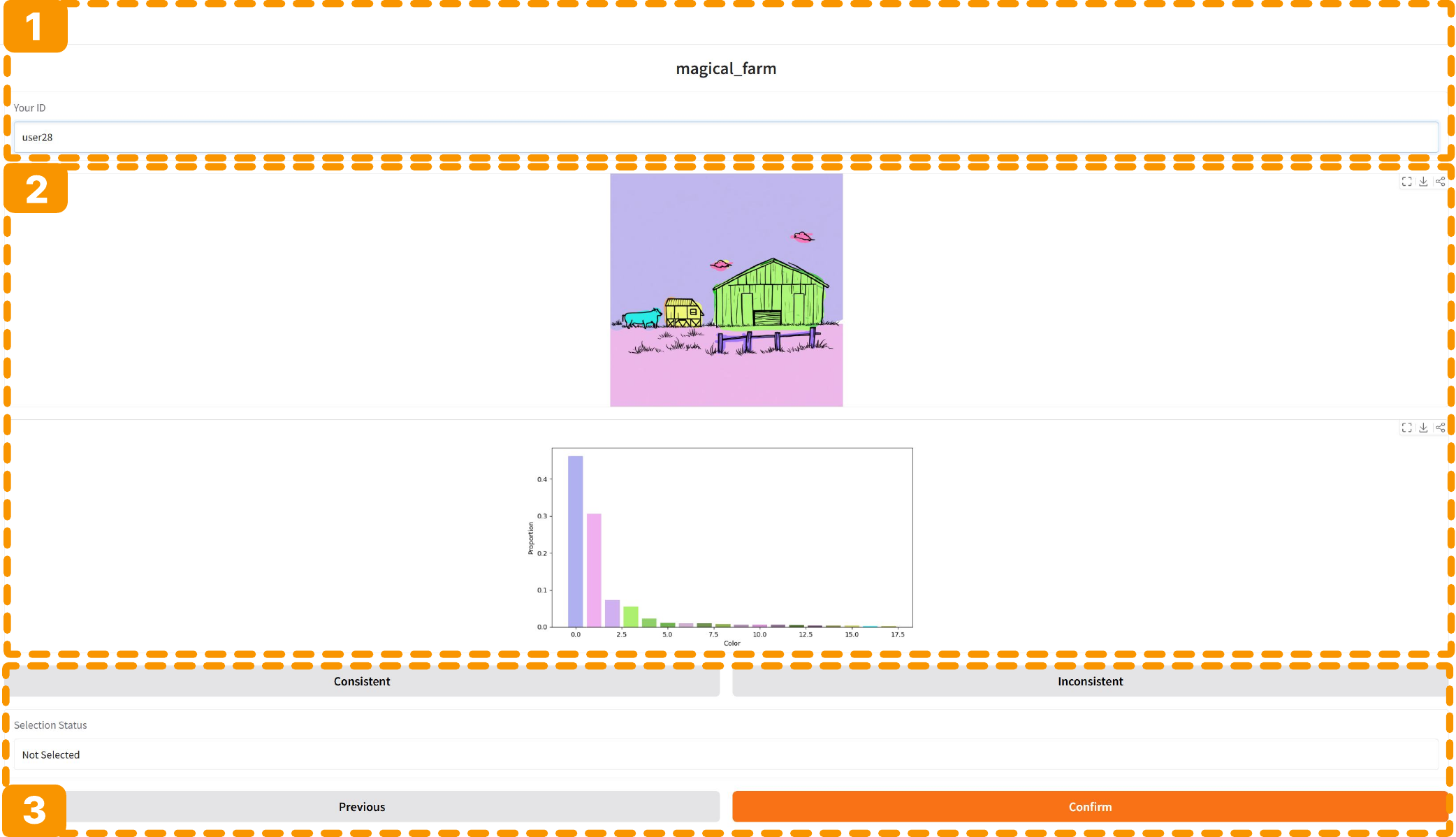}
    \caption{Quality control system interface.}
    \label{fig:systemqualitycontrol}
\end{subfigure}%
\hfill
\begin{subfigure}{0.45\textwidth}
    \centering
    \includegraphics[width=\linewidth]{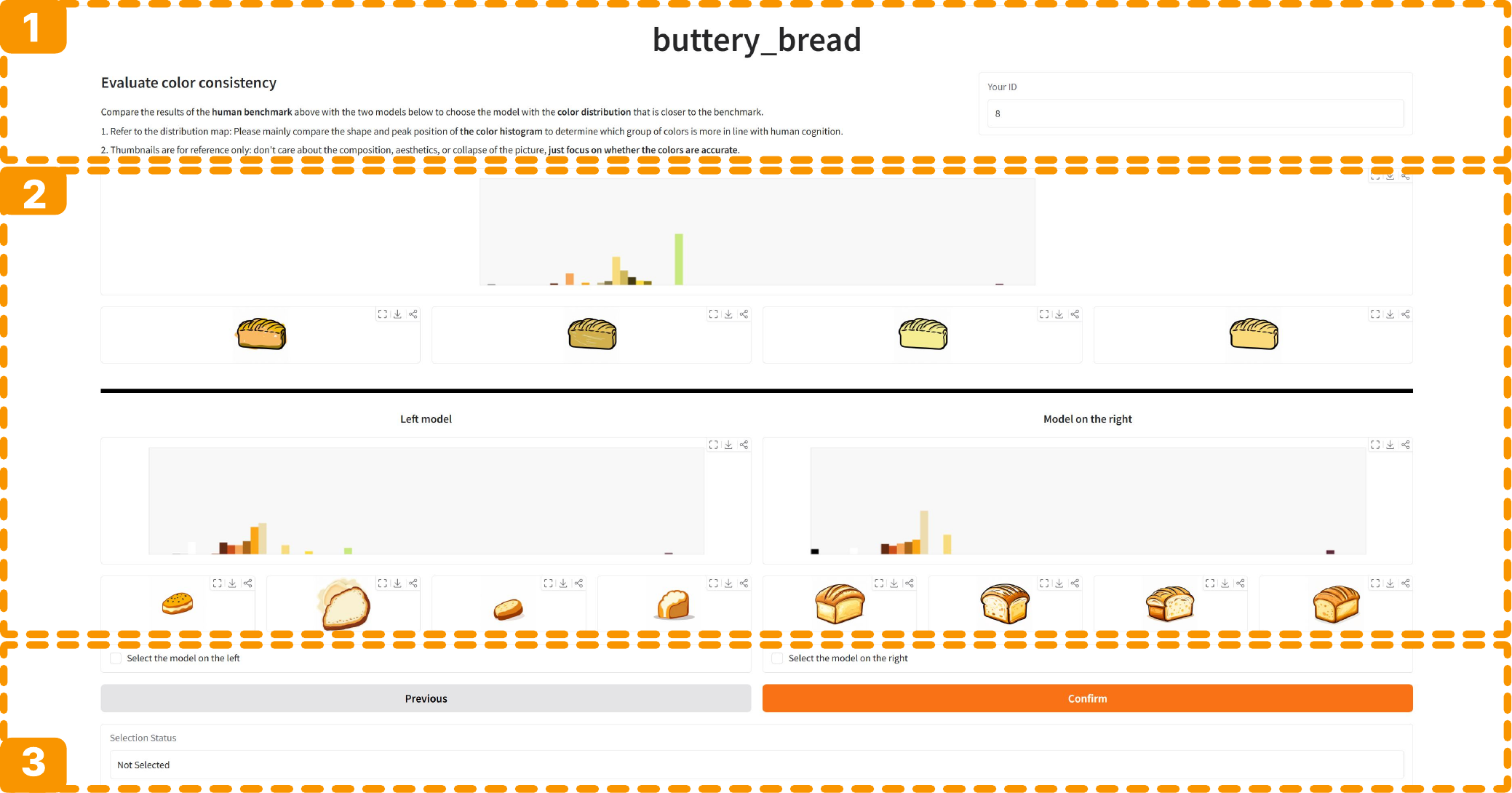}
    \caption{Human judgment interface.}
    \label{fig:appendix_humanrating}
\end{subfigure}
\caption{Gradio System for quality control and quantitative evaluation.
}
\label{fig:gradio}
\end{figure*}

% \begin{figure}[t]
%     \centering
%         \caption{Quantitative evaluation of sensitivity across different models with modifier categories.}
%     \includegraphics[width=0.98\linewidth]{fig/system-human.pdf}
%     \caption{Human judgment system interface.}
%     \label{fig:appendix_humanrating}
% \end{figure}

% \begin{figure}[htb]
%     \centering
%     \includegraphics[width=0.68\linewidth]{fig/system-qualitycontrol.pdf}
%     \caption{Quality control system interface}
%     \label{fig:systemqualitycontrol}
% \end{figure}
\begin{figure}[htb]
    \centering
    \includegraphics[width=0.48\linewidth]{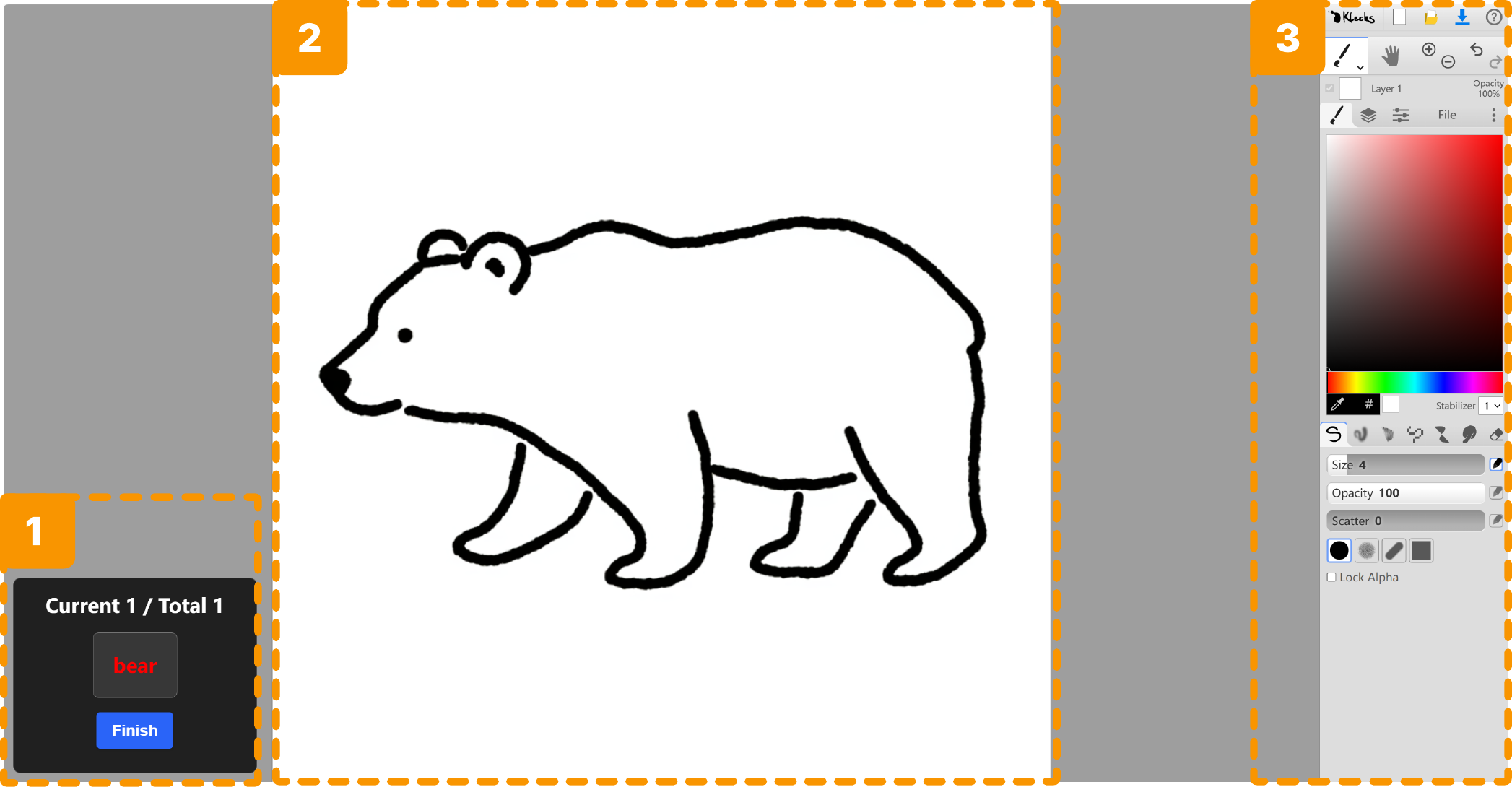}
    \caption{Annotation system interface}
    \label{fig:systemannotation}
\end{figure}

\begin{figure}[htb]
    \centering
    \includegraphics[width=0.78\linewidth]{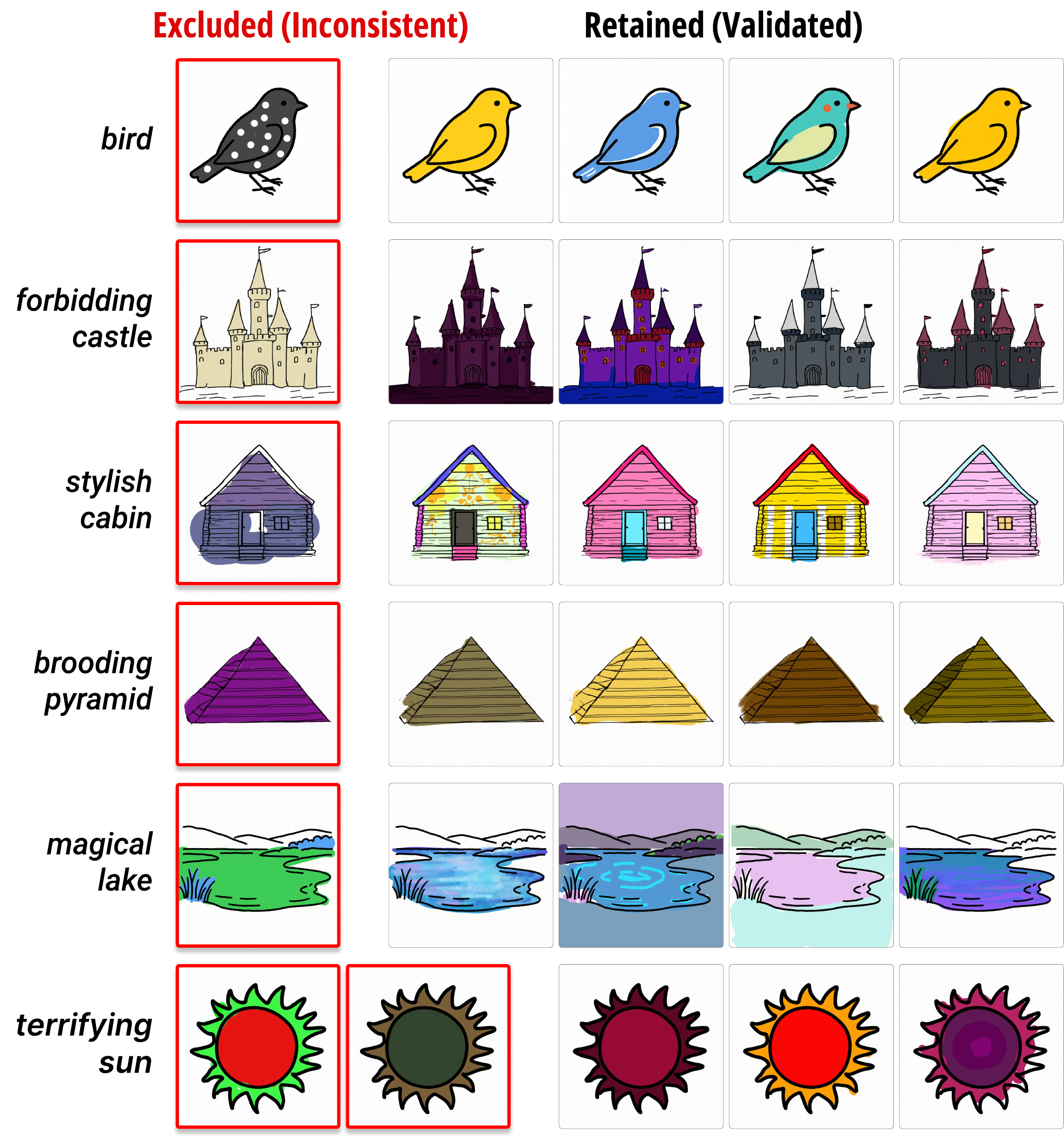}
    \caption{Examples of quality control for human-grounded color annotations. For concepts with high inter-annotator variance, we conducted a blind expert verification Y/N task. Samples marked in red (left column) were identified as semantically inconsistent outliers (e.g., a "forbidding castle" colored in bright pastels) and excluded. The retained instances (right columns) preserve the diverse but valid color distributions aligned with human cognition.}
    \label{fig:qualitycontrol}
\end{figure}
\paragraph{4) Inter-Expert Agreement Analysis.}
To quantify inter-expert agreement during the binary validation process, we further analyze the distribution of expert votes across all reviewed images.
Each image is independently evaluated by three experts and labeled as either Consistent or Inconsistent with respect to the semantic meaning of the concept.

As shown in Table~\ref{tab:expert_agreement}, the majority of images receive unanimous judgments (3/3), indicating a high level of consensus among experts.
A smaller portion of images exhibited partial disagreement (2 Yes / 1 No or 1 Yes / 2 No), suggesting variability in expert judgments for certain abstract concepts.
Images with unanimous or majority Inconsistent votes (3 No or 2 No) were discarded from the dataset to ensure the quality and semantic consistency of the final annotations.

Based on these annotations, we compute the average observed agreement across all reviewed images.
Following Fleiss' formulation, the observed agreement for each image $P_i$ is defined as the proportion 
of matching expert labels, and the overall agreement is obtained by averaging over all images:
\begin{equation}
\bar{P} = \frac{1}{N} \sum_{i=1}^{N} P_i ,
\end{equation}
where $N$ denotes the total number of reviewed images.

The resulting mean observed agreement is $\bar{P} = 0.811$, indicating a high level of consensus among 
experts despite the presence of semantically ambiguous cases.

\begingroup
\subsection{Experiment with Internet-Derived Ground Truth}
\label{sec:appendix_internet_derived}

To investigate when Internet images can provide a useful reference, we constructed Internet-derived color distributions for a randomly sampled subset of 100 concepts.

\paragraph{Data Collection and Filtering.}
For each concept, we collected the first 30 images returned by Google Images.
After deduplication, the collection contained 2,989 candidate images.
Eighteen participants screened the candidates and selected images whose visual appearance and colors were consistent with the corresponding concept.
An image was retained only if it was approved by at least three annotators.
This process retained 1,218 images and excluded 1,771 images, corresponding to a retention rate of 40.75\%.

\paragraph{Color Extraction and Evaluation.}
We subsequently applied the same target segmentation and color-extraction pipeline to the retained images. 
Among the 100 concepts, 96 yielded valid segmented samples and could therefore support an Internet-derived reference distribution. 
We evaluated the same models on these 96 concepts using both the human-derived and Internet-derived references.
Results for the natural and clipart styles are reported in Tables~\ref{tab_internet_natural} and~\ref{tab_internet_clipart}.
We also uploaded the final ground-truth histograms for the 96 concepts used in this experiment at \href{https://huggingface.co/datasets/ColorConceptBench/ColorConceptBench}{ColorConceptBench/Dataset\&Code}.

\begin{table}[t]
    \centering
    \caption{Natural-style performance.}
    \label{tab_internet_natural}
    \footnotesize
    \setlength{\tabcolsep}{3pt}
    \renewcommand{\arraystretch}{1.15}
    \begin{tabular}{clcccc}
        \toprule
        & & \multicolumn{2}{c}{EMD $\downarrow$} & \multicolumn{2}{c}{PCC $\uparrow$} \\
        \cmidrule(lr){3-4}
        \cmidrule(lr){5-6}
        Rank & Model & Human-derived Ref. & Internet-derived Ref. & Human-derived Ref. & Internet-derived Ref. \\
        \midrule
        1 & SD XL             & 28.507 [26.574, 30.441] & 18.652 [16.960, 20.344] & 0.341 [0.290, 0.392] & 0.637 [0.585, 0.689] \\
        2 & Flux              & 29.055 [26.903, 31.208] & 20.638 [18.956, 22.320] & 0.368 [0.316, 0.420] & 0.603 [0.551, 0.655] \\
        3 & Sana              & 29.192 [27.108, 31.276] & 21.232 [19.111, 23.354] & 0.372 [0.319, 0.425] & 0.597 [0.541, 0.652] \\
        4 & SD 3              & 30.394 [28.185, 32.603] & 21.306 [19.355, 23.257] & 0.320 [0.270, 0.369] & 0.568 [0.509, 0.626] \\
        5 & Pixart-$\alpha$    & 31.331 [28.986, 33.676] & 22.005 [19.922, 24.087] & 0.341 [0.289, 0.393] & 0.597 [0.543, 0.651] \\
        6 & Qwen-Image        & 31.460 [29.225, 33.696] & 22.172 [20.122, 24.223] & 0.313 [0.259, 0.366] & 0.580 [0.522, 0.637] \\
        7 & SD 3.5            & 33.177 [30.845, 35.509] & 22.727 [20.524, 24.930] & 0.283 [0.228, 0.339] & 0.548 [0.489, 0.607] \\
        8 & OmniGen2          & 32.938 [30.350, 35.526] & 24.559 [22.351, 26.768] & 0.327 [0.270, 0.385] & 0.530 [0.475, 0.585] \\
        9 & OmniGen           & 32.875 [30.423, 35.326] & 24.829 [22.507, 27.150] & 0.323 [0.271, 0.376] & 0.501 [0.441, 0.561] \\
        \bottomrule
    \end{tabular}
\end{table}

\begin{table}[t]
    \centering
    \caption{Clipart-style performance.}
    \label{tab_internet_clipart}
    \footnotesize
    \setlength{\tabcolsep}{3pt}
    \renewcommand{\arraystretch}{1.15}
    \begin{tabular}{clcccc}
        \toprule
        & & \multicolumn{2}{c}{EMD $\downarrow$} & \multicolumn{2}{c}{PCC $\uparrow$} \\
        \cmidrule(lr){3-4}
        \cmidrule(lr){5-6}
        Rank & Model & Human-derived Ref. & Internet-derived Ref. & Human-derived Ref. & Internet-derived Ref. \\
        \midrule
        1 & SD XL             & 27.689 [25.887, 29.491] & 21.403 [19.497, 23.309] & 0.351 [0.300, 0.402] & 0.570 [0.518, 0.623] \\
        2 & Pixart-$\alpha$    & 32.300 [29.919, 34.680] & 26.067 [23.580, 28.554] & 0.329 [0.278, 0.380] & 0.506 [0.450, 0.562] \\
        3 & Flux              & 29.429 [27.310, 31.547] & 26.255 [24.151, 28.359] & 0.342 [0.288, 0.396] & 0.464 [0.411, 0.517] \\
        4 & Sana              & 29.358 [26.997, 31.718] & 26.481 [23.674, 29.289] & 0.394 [0.346, 0.442] & 0.491 [0.437, 0.544] \\
        5 & SD 3.5            & 32.499 [30.558, 34.440] & 27.039 [24.677, 29.401] & 0.301 [0.246, 0.355] & 0.482 [0.429, 0.536] \\
        6 & SD 3              & 32.378 [30.202, 34.554] & 28.709 [26.037, 31.382] & 0.299 [0.247, 0.352] & 0.449 [0.392, 0.506] \\
        7 & Qwen-Image        & 31.989 [29.678, 34.299] & 29.312 [27.090, 31.534] & 0.284 [0.235, 0.334] & 0.412 [0.357, 0.466] \\
        8 & OmniGen           & 33.573 [31.053, 36.093] & 32.078 [29.425, 34.731] & 0.338 [0.284, 0.393] & 0.356 [0.296, 0.417] \\
        9 & OmniGen2          & 38.432 [35.519, 41.345] & 38.568 [35.710, 41.427] & 0.305 [0.245, 0.366] & 0.284 [0.230, 0.338] \\
        \bottomrule
    \end{tabular}
\end{table}

\paragraph{Consistent model rankings.}
The Internet-derived evaluation largely preserves the overall model ranking obtained with the human-derived reference, particularly for EMD. 
Across both styles, SD XL, Flux, and Sana remain among the strongest performers, while OmniGen and OmniGen2 remain near the lower end. 
Although some intermediate rankings differ under PCC, the main conclusions about relative model performance remain consistent.

\paragraph{Coverage of Common Concepts.}
Internet retrieval can efficiently expand coverage for common concepts with diverse real-world appearances. 
For example, ``candy'' has no single intrinsic color and may plausibly be red, yellow, green, or multicolored. 
Internet images can provide observations of these different modes, while achieving comparable coverage through designer coloring would require substantially more annotations. 
This advantage is particularly relevant to Natural-style evaluation, because fully capturing the color diversity present in natural images through designer coloring alone would require substantially more annotations.

\paragraph{Uneven Coverage of Modified Concepts.}
This advantage does not extend uniformly across the concept space. Only 40.75\% of the retrieved candidates passed human screening, and four concepts did not yield a valid reference distribution. 
Concepts containing modifiers such as ``charred,'' ``stale,'' ``shriveled,'' and ``rotten'' also frequently retained no more than five valid images.
In these cases, retrieved images were often irrelevant or expressed the modifier through contextual elements rather than through changes in the target object’s appearance. 
Thus, retrieval scale alone does not guarantee reliable distribution that is consistent with human interpretation, even when many candidate images are initially retrieved.

\paragraph{Discussion.}
This experiment shows both the usefulness and the limitations of Internet images. 
The close agreement in model rankings supports their value as an additional reference source. 
Their scale and natural diversity also make them promising for constructing training data, especially for common concepts with many valid examples. 
However, their uneven coverage of modified concepts makes them less suitable as the sole ground truth for our benchmark. 
We therefore view the two data sources as complementary. 
Human coloring provides a controlled and high-quality reference for evaluation, especially for visual-state or emotional modifiers. 
Carefully filtered and legally reusable Internet images can instead provide the scale and diversity for training. 
Future work could construct a hybrid dataset that uses human annotations as high-quality evaluation anchors and filtered Internet images to expand training coverage, with strictly separated training and evaluation subsets.
\vspace{-2mm}
% ==========================================
% SECTION B: IMPLEMENTATION (EXPERIMENT)
% ==========================================

\par\endgroup
\section{Implementation Details}
\label{sec:appendix_implementation}

\subsection{Model Inference Configuration}
\label{sec:appendix_model_inference}

To evaluate the model's capability in associating concepts with colors across different settings, we conduct a systematic image generation process. 
Image generation is performed using Nvidia A800 GPUs.
For each concept in our benchmark, we generate images under the following protocols:

\paragraph{1) Visual Style.} 
We explore \textit{natural photo} and \textit{clipart cartoon} as two common domains.
This choice enables us to evaluate if the model captures the colors of concepts universally, or if its performance is biased towards a specific visual style. 

\paragraph{2) Classifier-Free Guidance (CFG).} 
We focus on the CFG scale, a hyperparameter that controls the trade-off between alignment to the text prompt and image diversity. 
To investigate whether the guidance scale influences the model's color-concept association or the diversity of color selection, we evaluate the model across 7 distinct guidance scales. 
This aims to determine if the model's ability is robust to varying generation constraints.

\paragraph{3) Sampling Strategy.} 
For each unique combination of concept, style, and guidance scale, we generate 5 independent samples at a resolution of $1024 \times 1024$ with 50 inference steps, using distinct random seeds.

\subsection{Prompt Templates}
\label{sec:appendix_prompt_templates}
We utilize standardized prompt templates to trigger concept generation across different styles:
\begin{itemize}
    \item \textbf{Implicit Association (Ours):} ``A [Style] of a [Adjective] [Object], centered composition.'' (e.g., \textit{``A natural photo of a lonely street.''})
\end{itemize}

\subsection{Color Extraction Pipeline}
\label{sec:color_extraction}
\begin{figure}[htb]
    \centering
    \includegraphics[width=0.6\linewidth]{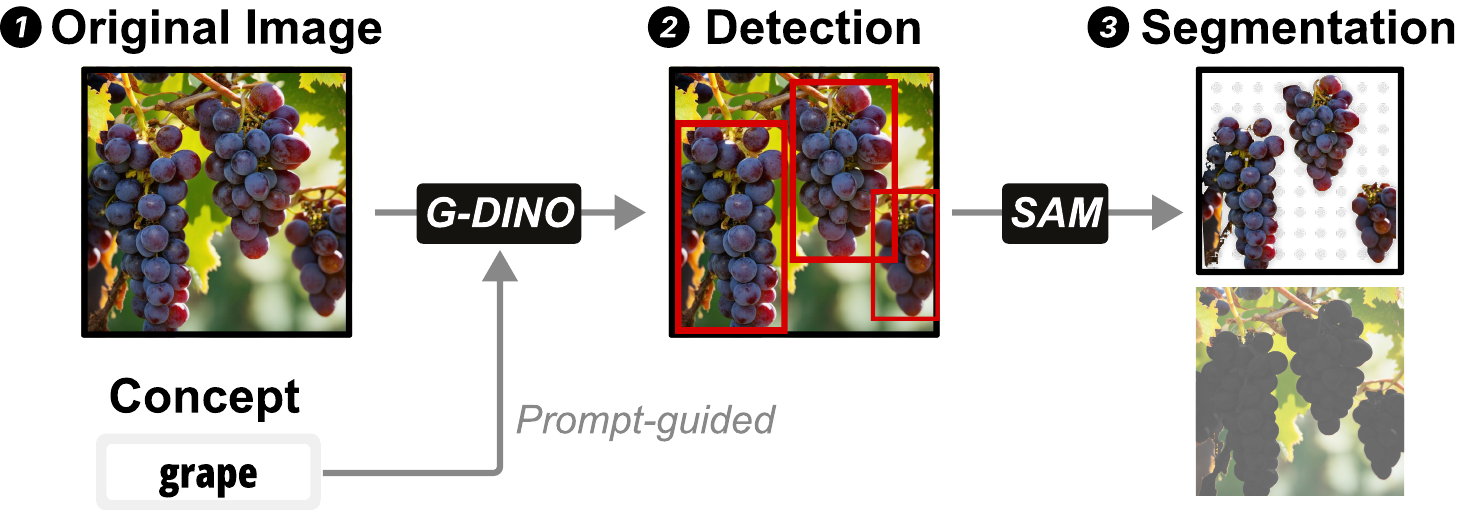}
    \caption{Segmentation pipeline.}
    \label{fig:appendix_segpipeline}
\end{figure}
To ensure that our analysis focuses exclusively on the target concept rather than the background environment, we employ a segmentation-based extraction pipeline \mbox{followed by adaptive color quantization.}

\paragraph{Segmentation.} 
\label{para:seg_filter}
\begin{figure}[htb]
    \centering
    \includegraphics[width=0.58\linewidth]{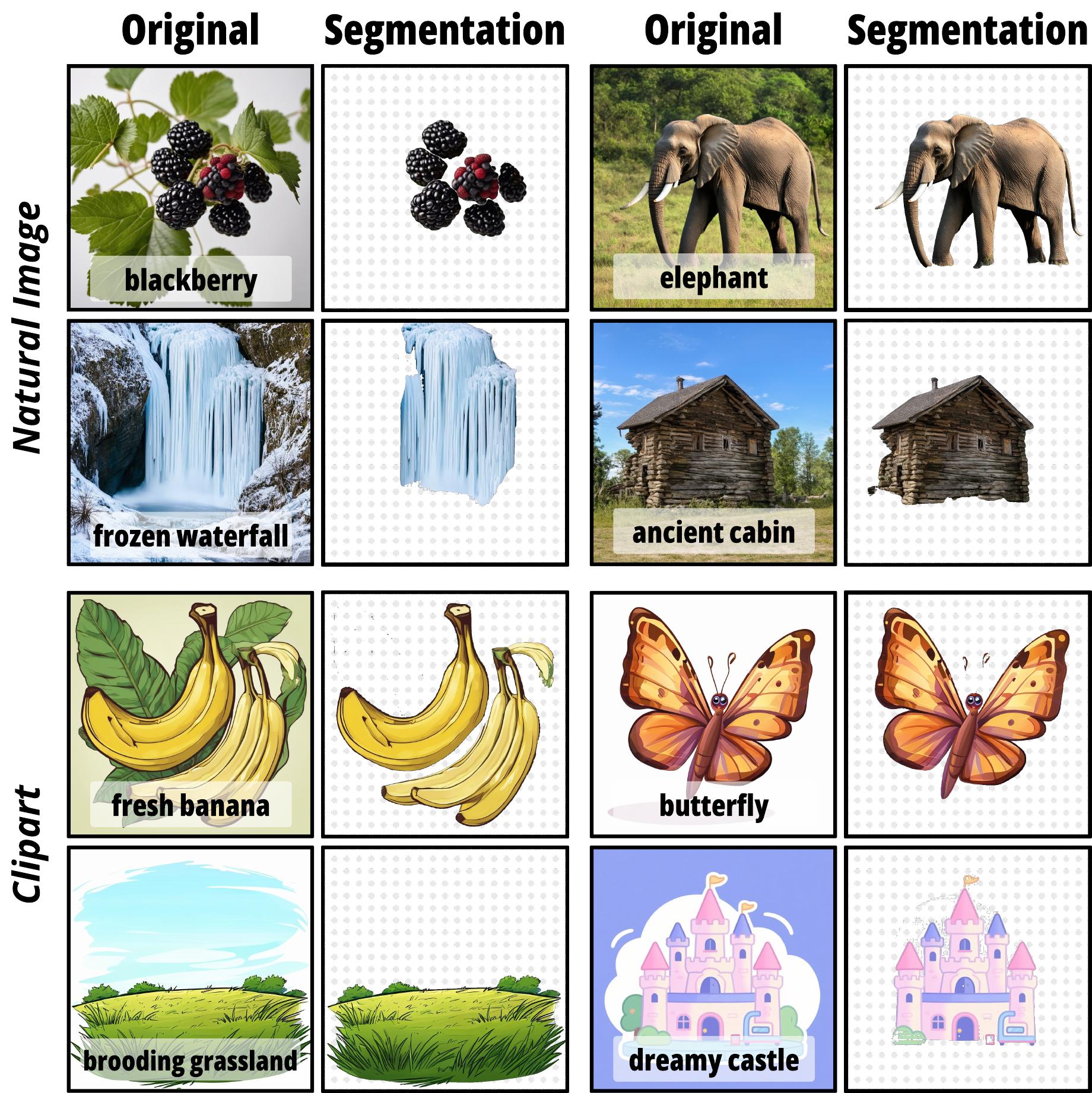}
    \caption{Color grounding using SAM.}
    \label{fig:appendix_seg}
\end{figure}
To extract concept-level color information, we apply a two-stage segmentation pipeline using Grounding DINO \citep{liu2024grounding} for bounding box detection followed by SAM \citep{kirillov2023segment} for binary mask generation.
To validate segmentation quality, we first compute the CLIPScore \citep{hessel2021clipscore} between the cropped segmented region and the concept name. 
For samples with a score below 0.15, we apply two additional geometric checks to confirm segmentation failure before removal: 
\begin{enumerate}
\item An area ratio check, comparing the size of the segmented region to the original image to detect masks that retain most of the background.
\item A boundary and center check, where masks that preserve pixels along all four edges of the image while failing to cover the center region are flagged as background captures.
\end{enumerate}
The segmentation pipeline achieves an initial success rate of 97.39\%, and we further perform manual verification on the remaining low-confidence samples.
Finally, 1,308 samples are removed from the final dataset.

\begingroup
\paragraph{Quantization and Refinement.} 
To capture the primary visual impression, we downsample the masked foreground to $100 \times 100$ pixels.
We divide the RGB space into $16 \times 16 \times 16$ bins and map filtered pixels into these bins.
We then implement an \textbf{adaptive grouping strategy}:
% \begin{enumerate}
%     \item We calculate the CIEDE2000 distance between the dominant colors of different images, clustering images into the same visual group if their dominant colors are perceptually similar ($\Delta E_{00} \le 12$).
%     \item Within each group, we aggregate pixels and quantize them using $8 \times 8 \times 8$ RGB bins.
%     \item The top 20 colors undergo a final merging process where color centers indistinguishable to the human eye ($\Delta E_{00} \le 7$) are combined \citep{szafir2017modeling}.
% \end{enumerate}
% The final distribution $q$ is constructed by aggregating these refined palettes, weighted by the number of images in each group.
\begin{enumerate}
    \item \textbf{Initial Quantization and Image Grouping.}
    We obtain a dominant color for each image from its quantized histogram and compare dominant colors in CIELAB using CIEDE2000.
    Images are organized into visual groups using a grouping threshold of $\Delta E_{00} \le 12$, allowing distinct color modes, such as red and green apples, to be processed separately.
    \item \textbf{Within-Group Palette Refinement.}
    Within each group, foreground pixels are pooled and quantized using an $8 \times 8 \times 8$ RGB grid.
    We retain the 20 most frequent color bins and merge their representative colors using a CIEDE2000 threshold of $\Delta E_{00} \le 7$.
    This refinement consolidates nearby colors into a compact palette.
    The refined group palettes are aggregated with weights proportional to the number of images in each group.
    The grouping and merging thresholds are fixed and applied uniformly across all concepts and models.
    \item \textbf{Projection onto UW71.}
    Each refined palette color is assigned to its nearest UW71 \citep{rathore2019estimating, hu2022self} prototype using CIEDE2000 distance.
    For each concept $c$, we accumulate the assigned weights and normalize them to construct the final human-grounded color distribution:
    \[
    P_{\text{H}}(\mathbf{u}_k \mid c)
    = \frac{m_k(c)}{\sum_{j=1}^{71} m_j(c)},
    \qquad k = 1,\ldots,71,
    \]
    where $\mathbf{u}_k$ denotes the $k$-th UW71 color prototype, and $m_k(c)$ is the total weight assigned to this prototype after weighting each visual group by its number of images.
    The same procedure is applied to model-generated images to construct $P_{\text{M}}(x \mid c)$.

\end{enumerate}

\par\endgroup
\begin{table}[htb]
    \centering
    \caption{Quantitative evaluation of sensitivity across different models with image style categories.}
    \begin{tabular}{lcc}
        \toprule
        \textbf{Method} & \multicolumn{2}{c}{\textbf{Style}}\\
        &\textbf{Clipart} & \textbf{Natural}  \\
        \midrule
        Flux   & 0.677 & 0.631 \\
        OmniGen  & 0.789 & 0.896 \\
        OmniGen2  & 1.027 & 0.800 \\
        Pixart-$\alpha$  & 0.742 & 0.688 \\
        Qwen-Image   & 0.861 & 1.047 \\
        Sana   & 0.765 & 0.592  \\
        SD 3    & 0.729 & 0.681  \\
        SD 3.5  & 0.810 & 0.686 \\
        SD XL   & 0.767 & 0.529  \\
        \bottomrule
    \end{tabular}

    \label{tab:sensitivity_scores_model}
\end{table}

\begin{table}[htb]
\caption{Quantitative evaluation of sensitivity across
different models with modifier categories.}
    \centering
    \begin{tabular}{lcc}
        \toprule
        \textbf{Method} & \multicolumn{2}{c}{\textbf{Modifier}}\\
        &\textbf{Visual State} & \textbf{Emotional}  \\
        \midrule
        Flux   & 0.656 & 0.645 \\
        OmniGen  & 0.860 & 0.729 \\
        OmniGen2  & 0.904 & 0.948 \\
        Pixart-$\alpha$  & 0.709 & 0.752 \\
        Qwen-Image   & 0.974 & 0.818 \\
        Sana   & 0.681 & 0.662  \\
        SD 3    & 0.711 & 0.668  \\
        SD 3.5  & 0.758 & 0.679 \\
        SD XL   & 0.665 & 0.536  \\
        \bottomrule
    \end{tabular}

    \label{tab:sensitivity_scores_model_mod}
\end{table}

\vspace{-2mm}
\section{Human Judgment}
\label{sec:appendix_metrics}
We conducted a pairwise human evaluation on clipart and natural images to validate the alignment between our metrics and human perception.
We sampled three models: Sana-1.5, OmniGen, and SDXL, representing a diverse range of EMD performance. 
The study utilized a Gradio interface where annotators compared two model outputs against a human ground truth.
Our system employs a non-repeating random assignment mechanism to distribute tasks and randomly place the candidate models to mitigate positional bias.
The study involved 62 participants (24 with design/art backgrounds and 38 non-experts).
Each case received 5 independent votes, yielding a total of 4,350 valid responses.

\textbf{Annotation Instructions.} 
To ensure evaluations focused strictly on color semantics rather than image quality, we provide following instructions: (1) Participants are directed to judge based on both the model-generated images and their extracted color histograms as complementary references. (2) We explicitly instruct annotators to disregard composition, aesthetic appeal, or generation artifacts. The core task defined as ``selecting the model whose color distribution best matches the human ground-truth.''

\begin{table}[htb]
    \centering
    \small
    \caption{Quantitative evaluation of sensitivity across different categories.}
    \begin{tabular}{lccc}
        \toprule
        \textbf{Method} & \multicolumn{2}{c}{\textbf{Style}}\\
        &\textbf{Clipart} & \textbf{Natural}&\textbf{Average}  \\
        \midrule
        Emotional    & 0.773 & 0.659 & 0.716 \\
        Visual State & 0.800 & 0.739 & 0.769 \\
        \bottomrule
    \end{tabular}

    \label{tab:sensitivity_scores_cate}
\end{table}   
\vspace{-2mm}
\section{Deterministic Metric}
\label{sec:appendix_deterministic}
Similar to previous works relying on deterministic metrics, we further extract dominant color from the underlying distributions to evaluate features alignments.

\paragraph{Dominant Color Accuracy (DCA).}
We identify the ``dominant color'' as the bin with the highest probability mass in the aggregated distribution.
For a specific concept $c$, let  $k_H^{(c)} = \arg\max_{k} (p_k)$ and $k_M^{(c)} = \arg\max_{k} (q_k)$ be the indices of the peak color bins for humans and the model.
We calculate the accuracy over the concept set $\mathcal{C}$:

\begin{equation}
\small
DCA = \frac{1}{|\mathcal{C}|} \sum_{c \in \mathcal{C}} \mathds{1}(k_H^{(c)} = k_M^{(c)}),
\end{equation}

where $k_H^{(c)}$ and $k_M^{(c)}$ denote the dominant colors, and $\mathds{1}(\cdot)$ is the indicator function.
This metric strictly assesses the model's ability to precisely capture the representative color of the concept.

\paragraph{Hue Angular Difference ($\Delta \text{Hue}$).}
To explicitly evaluate chromatic alignment, we compute the angular difference between the mean hue angles of the dominant color. 
For a concept $c$, let $\theta_H^{(c)}$ and $\theta_M^{(c)}$ be the hue angles associated with the dominant bins $k_H^{(c)}$ and $k_M^{(c)}$. 
The metric is defined as the shortest angular distance between these two angles, averaged over all concepts:

\begin{equation}
\small
\Delta Hue = \frac{1}{|\mathcal{C}|} \sum_{c \in \mathcal{C}} \min(|\Delta \theta^{(c)}|, 360^\circ - |\Delta \theta^{(c)}|),
\vspace{-2mm}
\end{equation}

where $\Delta \theta^{(c)} = \theta_H^{(c)} - \theta_M^{(c)}$.
This metric quantifies the divergence in the overall hue direction.

\begingroup
\paragraph{Deterministic Feature Alignment.}
Table~\ref{tab:dominant_metric} presents the deterministic feature alignment of T2I models across concept categories and visual styles.
For DCA, Original concepts generally achieve the highest scores, while performance on Visual State and Emotional variants varies across models and settings, consistent with the variability observed in the probabilistic metrics.
Most models achieve relatively low $\Delta$Hue on Original Clipart prompts, errors increase substantially for Emotional concepts, with several models exceeding 50° in the Natural style.
This suggests that abstract semantic modifiers can affect both distributional alignment and dominant hue fidelity, as models do not consistently capture the shift in dominant color that emotional modifiers are expected to induce.
Across models, Flux.1-dev and Sana-1.5 achieve the strongest DCA scores, while OmniGen2 and SD~3 record the lowest $\Delta$Hue errors in select conditions, suggesting that hue precision and distributional alignment do not always co-vary across models.

\par\endgroup
\begin{table*}   
\caption{Comparison of \textit{deterministic feature alignment} of different T2I models across concepts and styles. The best and second best results in each column are marked in \textbf{bold} and \underline{underlined}, respectively.}
\scriptsize    
\centering    
\setlength{\tabcolsep}{1.2pt}    
\begin{tabular}{c cc cc cc cc cc cc}        
\toprule        
\multirow{3}{*}{\textbf{Method}} & \multicolumn{4}{c}{\textbf{Original}} & \multicolumn{4}{c}{\textbf{Visual state}} & \multicolumn{4}{c}{\textbf{Emotional}}\\                                         & \multicolumn{2}{c}{\textbf{Natural}}  & \multicolumn{2}{c}{\textbf{Clipart}}      & \multicolumn{2}{c}{\textbf{Natural}}   & \multicolumn{2}{c}{\textbf{Clipart}} & \multicolumn{2}{c}{\textbf{Natural}} & \multicolumn{2}{c}{\textbf{Clipart}}\\        
\cmidrule(lr){2-3}\cmidrule(lr){4-5}\cmidrule(lr){6-7}\cmidrule(lr){8-9}\cmidrule(lr){10-11}\cmidrule(lr){12-13}                                         & DCA$\uparrow$ & $\Delta$ Hue$\downarrow$ & DCA$\uparrow$ & $\Delta$ Hue$\downarrow$ & DCA$\uparrow$ & $\Delta$ Hue$\downarrow$ & DCA$\uparrow$ & $\Delta$ Hue$\downarrow$ & DCA$\uparrow$ & $\Delta$ Hue$\downarrow$ & DCA$\uparrow$ & $\Delta$ Hue$\downarrow$ \\        
\midrule        
Flux.1-dev & \textbf{0.187} & 33.952 & \textbf{0.201} & \textbf{27.987} & \textbf{0.172} & \underline{38.647} & \underline{0.170} & 36.656 & \underline{0.233} & 45.881 & 0.120 & 58.490 \\
OmniGen & \underline{0.178} & 35.295 & 0.168 & 32.443 & 0.140 & 41.013 & 0.117 & 41.649 & 0.180 & 53.326 & 0.053 & 48.599 \\
OmniGen2 & 0.173 & \textbf{31.798} & 0.178 & 30.412 & 0.147 & 38.903 & 0.109 & \underline{35.708} & 0.135 & 53.816 & 0.098 & 50.383 \\        
PixArt-$\alpha$ & \underline{0.178} & 32.890 & 0.164 & 32.512 & 0.165 & 42.138 & 0.155 & 38.120 & \underline{0.233} & \underline{42.620} & \underline{0.203} & \underline{44.837} \\
Qwen-Image & 0.159 & 34.344 & 0.164 & 31.002 & 0.133 & 41.738 & 0.123 & \textbf{34.133} & 0.150 & 47.374 & 0.105 & 51.125 \\        
Stable Diffusion 3 & 0.173 & \underline{32.774} & 0.182 & 29.026 & 0.125 & 39.538 & 0.113 & 38.007 & 0.150 & \textbf{41.603} & 0.143 & \textbf{43.865} \\        
Stable Diffusion 3.5 & 0.159 & 33.270 & 0.168 & 33.073 & 0.124 & \textbf{37.743} & 0.123 & 37.099 & 0.105 & 48.782 & 0.135 & 44.859 \\        
Stable Diffusion XL & 0.084 & 39.101 & \underline{0.192} & 32.188 & 0.121 & 46.496 & 0.138 & 37.890 & 0.143 & 51.856 & 0.143 & 50.190 \\        
Sana-1.5 & 0.145 & 33.047 & 0.173 & \underline{28.647} & \underline{0.166} & 39.281 & \textbf{0.194} & 36.801 & \textbf{0.271} & 43.133 & \textbf{0.226} & 47.814 \\       
\bottomrule   
\end{tabular}  
\label{tab:dominant_metric}
\end{table*}
\vspace{-2mm}
% ==========================================
% SECTION E: VISUALIZATION
% ==========================================
\section{Additional Qualitative Results}
\label{sec:appendix_qualitative}
\begin{figure}[hb]
    \centering
    \includegraphics[width=0.50\linewidth]{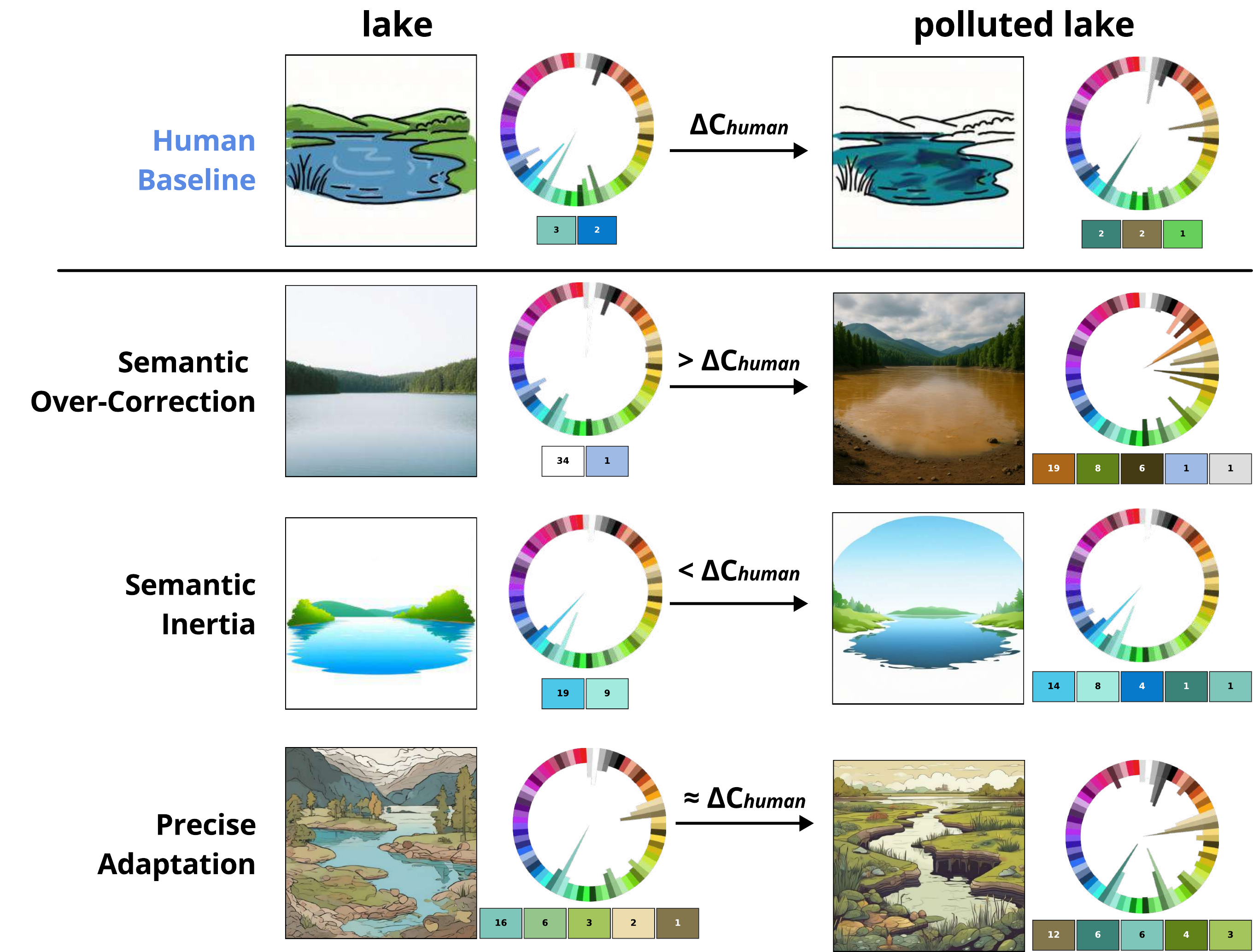}
    \caption{Qualitative results. Compared to the Human baseline (Top), models exhibit three distinct behaviors: Over-Correction (excessive shift, loss of identity), Inertia (negligible shift, failure to adapt), and Precise Adaptation (balanced shift, preserving object identity while updating attributes).}
    \label{fig:extra_case}
\end{figure}

\vspace{-2mm}
% ==========================================
% SECTION D: ADDITIONAL RESULTS
% ==========================================
\section{Additional Quantitative Results}
\label{sec:appendix_results}

\subsection{Modifier Sensitivity}
We present the quantitative evaluation of sensitivity in Table \ref{tab:sensitivity_scores_model}, Table \ref{tab:sensitivity_scores_model_mod}, and Table \ref{tab:sensitivity_scores_cate}. 
Table \ref{tab:sensitivity_scores_model} and Table \ref{tab:sensitivity_scores_model_mod} details the performance of individual models across different styles and different modifiers.
To provide a broader perspective, Table \ref{tab:sensitivity_scores_cate} summarizes the results by semantic modifier category (Emotional vs. Visual State), including the calculated average sensitivity scores \mbox{to facilitate an overall comparison.}

\begingroup
\subsection{Matched Comparison of Original and Modified Concepts}
\label{app_matched_comparison}
We add a matched analysis restricted to base concepts with Emotional or Visual variants. For each base concept, we average its variant scores and compare them with its Original score, defining $\Delta$ as adjective minus Original.
Tables~\ref{tab_matched_visual} and~\ref{tab_matched_emotional} report the results.

For Visual State variants, eight of nine models show worse alignment on both metrics for clipart images, whereas five improve on both metrics for natural images.
For Emotional variants, natural images achieve lower EMD for seven models and higher PCC for all nine, while clipart images exhibit mixed results.

\begin{table}[htbp]
    \centering
    \caption{Matched Original-to-Visual State.}
    \label{tab_matched_visual}
    \small
    \setlength{\tabcolsep}{8pt}
    \begin{tabular}{lrrrr}
        \toprule
        & \multicolumn{2}{c}{Clipart} & \multicolumn{2}{c}{Natural} \\
        \cmidrule(lr){2-3}
        \cmidrule(lr){4-5}
        Model & $\Delta\mathrm{EMD}$ & $\Delta\mathrm{PCC}$ & $\Delta\mathrm{EMD}$ & $\Delta\mathrm{PCC}$ \\
        \midrule
        Flux             &  2.423 & -0.057 & -2.081 &  0.022 \\
        OmniGen2         &  4.232 & -0.100 &  1.978 & -0.045 \\
        OmniGen          &  3.759 & -0.083 &  0.847 & -0.011 \\
        Pixart-$\alpha$   &  2.580 & -0.057 & -2.076 &  0.018 \\
        Qwen-Image       &  1.118 & -0.059 & -0.763 &  0.012 \\
        Sana             &  2.389 & -0.043 & -2.691 &  0.052 \\
        SD 3.5           &  0.971 & -0.023 &  1.197 & -0.013 \\
        SD 3             &  3.319 & -0.075 &  0.074 & -0.001 \\
        SD XL            & -1.009 &  0.033 & -3.539 &  0.073 \\
        \bottomrule
    \end{tabular}
\end{table}

\begin{table}[htbp]
    \centering
    \caption{Matched Original-to-Emotional.}
    \label{tab_matched_emotional}
    \small
    \setlength{\tabcolsep}{8pt}
    \begin{tabular}{lrrrr}
        \toprule
        & \multicolumn{2}{c}{Clipart} & \multicolumn{2}{c}{Natural} \\
        \cmidrule(lr){2-3}
        \cmidrule(lr){4-5}
        Model & $\Delta\mathrm{EMD}$ & $\Delta\mathrm{PCC}$ & $\Delta\mathrm{EMD}$ & $\Delta\mathrm{PCC}$ \\
        \midrule
        Flux             &  0.276 &  0.025 & -3.413 & 0.089 \\
        OmniGen2         &  3.444 & -0.024 & -0.241 & 0.032 \\
        OmniGen          &  2.436 &  0.021 & -4.716 & 0.105 \\
        Pixart-$\alpha$   & -1.112 &  0.099 & -5.802 & 0.114 \\
        Qwen-Image       &  1.424 &  0.017 &  0.283 & 0.064 \\
        Sana             & -2.995 &  0.082 & -4.027 & 0.072 \\
        SD 3.5           &  0.882 & -0.008 &  0.417 & 0.015 \\
        SD 3             &  1.372 & -0.016 & -3.777 & 0.109 \\
        SD XL            & -0.530 &  0.050 & -3.650 & 0.091 \\
        \bottomrule
    \end{tabular}
\end{table}

Table~\ref{tab_fig5_quartiles} reports the quartile statistics underlying Figure~5.
Tables~\ref{tab_fig6_emd} and~\ref{tab_fig6_pcc} report the numerical results and confidence intervals underlying Figure~6.

\begin{table}[htbp]
    \centering
    \caption{Quartile statistics corresponding to the Overall boxplots in Figure~5.}
    \label{tab_fig5_quartiles}
    \small
    \setlength{\tabcolsep}{12pt}
    \renewcommand{\arraystretch}{1.15}
    \begin{tabular}{lrrr}
        \toprule
        Method & Q1 & Median & Q3 \\
        \midrule
        Flux             & 0.346 & 0.542 & 0.855 \\
        OmniGen          & 0.446 & 0.720 & 1.108 \\
        OmniGen2         & 0.505 & 0.781 & 1.210 \\
        Pixart-$\alpha$   & 0.380 & 0.614 & 0.938 \\
        Qwen-Image       & 0.551 & 0.856 & 1.237 \\
        SD 3             & 0.393 & 0.612 & 0.897 \\
        SD 3.5           & 0.412 & 0.653 & 0.972 \\
        SD XL            & 0.364 & 0.551 & 0.829 \\
        Sana             & 0.363 & 0.565 & 0.883 \\
        \bottomrule
    \end{tabular}
\end{table}

\begin{table}[htbp]
    \centering
    \caption{EMD statistics corresponding to the guidance-scale trends in Figure~6, reported as mean $\pm$ 95\% CI half-width.}
    \label{tab_fig6_emd}
    \scriptsize
    \setlength{\tabcolsep}{2pt}
    \renewcommand{\arraystretch}{1.15}
    \begin{tabular}{lccccccc}
        \toprule
        & \multicolumn{7}{c}{Guidance Scale} \\
        \cmidrule(lr){2-8}
        Method & 2 & 3 & 4 & 5 & 6 & 7 & 8 \\
        \midrule
        Flux           & $27.569 \pm 0.456$ & $28.059 \pm 0.454$ & $28.499 \pm 0.455$ & $28.685 \pm 0.461$ & $28.790 \pm 0.452$ & $29.422 \pm 0.462$ & $30.160 \pm 0.477$ \\
        OmniGen        & $31.181 \pm 0.504$ & $33.476 \pm 0.540$ & $34.809 \pm 0.562$ & $35.535 \pm 0.564$ & $36.418 \pm 0.582$ & $36.826 \pm 0.577$ & $37.843 \pm 0.607$ \\
        OmniGen2       & $37.464 \pm 0.616$ & $38.434 \pm 0.637$ & $39.459 \pm 0.652$ & $39.456 \pm 0.657$ & $39.151 \pm 0.657$ & $39.015 \pm 0.647$ & $39.114 \pm 0.658$ \\
        Pixart-$\alpha$ & $30.053 \pm 0.504$ & $30.552 \pm 0.503$ & $30.494 \pm 0.501$ & $30.294 \pm 0.503$ & $30.131 \pm 0.493$ & $30.173 \pm 0.493$ & $30.084 \pm 0.488$ \\
        Qwen-Image     & $31.181 \pm 0.453$ & $31.554 \pm 0.471$ & $31.701 \pm 0.479$ & $31.239 \pm 0.465$ & $30.986 \pm 0.464$ & $30.862 \pm 0.461$ & $30.750 \pm 0.474$ \\
        SD 3           & $29.387 \pm 0.473$ & $30.362 \pm 0.484$ & $30.823 \pm 0.491$ & $31.251 \pm 0.500$ & $31.403 \pm 0.500$ & $31.188 \pm 0.495$ & $31.478 \pm 0.504$ \\
        SD 3.5         & $29.811 \pm 0.442$ & $30.905 \pm 0.460$ & $31.255 \pm 0.469$ & $31.370 \pm 0.467$ & $31.408 \pm 0.470$ & $31.514 \pm 0.480$ & $31.401 \pm 0.470$ \\
        SD XL          & $27.656 \pm 0.442$ & $27.173 \pm 0.433$ & $27.256 \pm 0.420$ & $27.432 \pm 0.421$ & $27.571 \pm 0.424$ & $27.982 \pm 0.433$ & $28.287 \pm 0.441$ \\
        Sana           & $27.835 \pm 0.472$ & $28.077 \pm 0.478$ & $27.819 \pm 0.467$ & $27.870 \pm 0.463$ & $27.797 \pm 0.468$ & $27.708 \pm 0.458$ & $27.690 \pm 0.467$ \\
        \bottomrule
    \end{tabular}
\end{table}

\begin{table}[htp]
    \centering
    \caption{PCC statistics corresponding to the guidance-scale trends in Figure~6, reported as mean $\pm$ 95\% CI half-width.}
    \label{tab_fig6_pcc}
    \scriptsize
    \setlength{\tabcolsep}{2pt}
    \renewcommand{\arraystretch}{1.15}
    \begin{tabular}{lccccccc}
        \toprule
        & \multicolumn{7}{c}{Guidance Scale} \\
        \cmidrule(lr){2-8}
        Method & 2 & 3 & 4 & 5 & 6 & 7 & 8 \\
        \midrule
        Flux           & $0.371 \pm 0.028$ & $0.370 \pm 0.027$ & $0.370 \pm 0.027$ & $0.372 \pm 0.028$ & $0.372 \pm 0.028$ & $0.370 \pm 0.028$ & $0.370 \pm 0.028$ \\
        OmniGen        & $0.323 \pm 0.035$ & $0.323 \pm 0.035$ & $0.321 \pm 0.035$ & $0.323 \pm 0.035$ & $0.322 \pm 0.035$ & $0.324 \pm 0.035$ & $0.323 \pm 0.035$ \\
        OmniGen2       & $0.305 \pm 0.054$ & $0.305 \pm 0.054$ & $0.306 \pm 0.054$ & $0.304 \pm 0.053$ & $0.304 \pm 0.054$ & $0.305 \pm 0.054$ & $0.305 \pm 0.054$ \\
        Pixart-$\alpha$ & $0.369 \pm 0.035$ & $0.369 \pm 0.035$ & $0.369 \pm 0.034$ & $0.370 \pm 0.034$ & $0.370 \pm 0.035$ & $0.369 \pm 0.034$ & $0.369 \pm 0.034$ \\
        Qwen-Image     & $0.318 \pm 0.038$ & $0.318 \pm 0.038$ & $0.317 \pm 0.036$ & $0.318 \pm 0.038$ & $0.318 \pm 0.038$ & $0.318 \pm 0.038$ & $0.317 \pm 0.036$ \\
        SD 3           & $0.332 \pm 0.040$ & $0.332 \pm 0.040$ & $0.332 \pm 0.040$ & $0.332 \pm 0.040$ & $0.332 \pm 0.039$ & $0.332 \pm 0.040$ & $0.332 \pm 0.039$ \\
        SD 3.5         & $0.310 \pm 0.019$ & $0.311 \pm 0.019$ & $0.310 \pm 0.020$ & $0.310 \pm 0.019$ & $0.310 \pm 0.019$ & $0.311 \pm 0.019$ & $0.310 \pm 0.020$ \\
        SD XL          & $0.352 \pm 0.025$ & $0.352 \pm 0.026$ & $0.351 \pm 0.026$ & $0.352 \pm 0.025$ & $0.351 \pm 0.026$ & $0.351 \pm 0.026$ & $0.352 \pm 0.026$ \\
        Sana           & $0.404 \pm 0.033$ & $0.404 \pm 0.033$ & $0.404 \pm 0.033$ & $0.404 \pm 0.033$ & $0.404 \pm 0.033$ & $0.403 \pm 0.033$ & $0.404 \pm 0.033$ \\
        \bottomrule
    \end{tabular}
\end{table}

\subsection{Pairwise Significance Tests.}
We compare all 36 model pairs per style and metric using two-sided Wilcoxon signed-rank tests over paired concept-level scores, with Holm correction.
The analysis pools all prompt types, comprising 1,271 natural-style and 1,274 clipart-style concept prompts.
At $\alpha=0.05$, 117 of the 144 comparisons are significant: 34/36 for clipart EMD, 29/36 for natural EMD, and 27/36 for PCC in each style.
SD XL, Sana, and Flux each perform significantly better than OmniGen and OmniGen2 across both styles and metrics.
On the smaller Original subset of 216 concepts, significant differences primarily remain between leading and trailing models.
\vspace{-2mm}
\begin{table}[htbp]
    \centering
    \caption{
    Significant pairs are identified using two-sided Wilcoxon signed-rank tests with Holm correction at $\alpha=0.05$.
    }
    \label{tab_friedman_pairwise}
    \small
    \setlength{\tabcolsep}{8pt}
    \renewcommand{\arraystretch}{1.15}
    \begin{tabular}{llrrcc}
        \toprule
        Style & Metric & $n$ & $\chi^2(8)$ & Friedman $p$ & Significant pairs \\
        \midrule
        Natural & EMD & 1271 &  559.17 & $1.39 \times 10^{-115}$ & 29/36 \\
        Natural & PCC & 1271 &  300.07 & $3.98 \times 10^{-60}$  & 27/36 \\
        Clipart & EMD & 1274 & 1737.65 & $<10^{-300}$           & 34/36 \\
        Clipart & PCC & 1274 &  717.90 & $1.00 \times 10^{-149}$ & 27/36 \\
        \bottomrule
    \end{tabular}
\end{table}

\begin{table}[htbp]
    \centering
    \caption{Performance on Original concepts, reported as mean $\pm$ 95\% CI half-width.}
    \label{tab_original_ci}
    \small
    \setlength{\tabcolsep}{8pt}
    \renewcommand{\arraystretch}{1.15}
    \begin{tabular}{lcccc}
        \toprule
        & \multicolumn{2}{c}{Clipart} & \multicolumn{2}{c}{Natural} \\
        \cmidrule(lr){2-3}
        \cmidrule(lr){4-5}
        Model & EMD $\downarrow$ & PCC $\uparrow$ & EMD $\downarrow$ & PCC $\uparrow$ \\
        \midrule
        Flux             & $26.915 \pm 1.312$ & $0.391 \pm 0.036$ & $29.165 \pm 1.454$ & $0.369 \pm 0.034$ \\
        OmniGen          & $31.413 \pm 1.649$ & $0.363 \pm 0.035$ & $30.866 \pm 1.618$ & $0.348 \pm 0.035$ \\
        OmniGen2         & $36.349 \pm 1.806$ & $0.321 \pm 0.038$ & $30.066 \pm 1.624$ & $0.355 \pm 0.036$ \\
        Pixart-$\alpha$   & $29.663 \pm 1.556$ & $0.360 \pm 0.034$ & $30.867 \pm 1.633$ & $0.351 \pm 0.035$ \\
        Qwen-Image       & $30.577 \pm 1.489$ & $0.326 \pm 0.036$ & $30.311 \pm 1.564$ & $0.316 \pm 0.037$ \\
        Sana             & $26.151 \pm 1.353$ & $0.428 \pm 0.034$ & $29.425 \pm 1.442$ & $0.360 \pm 0.033$ \\
        SD 3             & $29.707 \pm 1.487$ & $0.359 \pm 0.037$ & $28.839 \pm 1.501$ & $0.348 \pm 0.035$ \\
        SD 3.5           & $30.206 \pm 1.332$ & $0.325 \pm 0.037$ & $29.442 \pm 1.450$ & $0.321 \pm 0.036$ \\
        SD XL            & $27.191 \pm 1.350$ & $0.344 \pm 0.036$ & $29.273 \pm 1.415$ & $0.308 \pm 0.033$ \\
        \bottomrule
    \end{tabular}
\end{table}

\subsection{Confidence Intervals by Concept Type.}
Tables~\ref{tab_original_ci},~\ref{tab_visual_ci}, and~\ref{tab_emotional_ci} report mean scores and 95\% confidence intervals for Original, Visual State, and Emotional concepts, respectively.

\begin{table}[htbp]
    \centering
    \caption{Performance on Visual State concepts, reported as mean $\pm$ 95\% CI half-width.}
    \label{tab_visual_ci}
    \small
    \setlength{\tabcolsep}{8pt}
    \renewcommand{\arraystretch}{1.15}
    \begin{tabular}{lcccc}
        \toprule
        & \multicolumn{2}{c}{Clipart} & \multicolumn{2}{c}{Natural} \\
        \cmidrule(lr){2-3}
        \cmidrule(lr){4-5}
        Model & EMD $\downarrow$ & PCC $\uparrow$ & EMD $\downarrow$ & PCC $\uparrow$ \\
        \midrule
        Flux             & $28.944 \pm 0.642$ & $0.337 \pm 0.017$ & $28.413 \pm 0.703$ & $0.378 \pm 0.016$ \\
        OmniGen          & $33.840 \pm 0.778$ & $0.278 \pm 0.016$ & $32.658 \pm 0.812$ & $0.319 \pm 0.017$ \\
        OmniGen2         & $39.130 \pm 0.852$ & $0.224 \pm 0.016$ & $31.572 \pm 0.746$ & $0.320 \pm 0.017$ \\
        Pixart-$\alpha$   & $30.554 \pm 0.744$ & $0.333 \pm 0.017$ & $29.747 \pm 0.733$ & $0.360 \pm 0.017$ \\
        Qwen-Image       & $31.676 \pm 0.684$ & $0.280 \pm 0.016$ & $30.600 \pm 0.685$ & $0.319 \pm 0.016$ \\
        Sana             & $28.243 \pm 0.679$ & $0.375 \pm 0.017$ & $27.861 \pm 0.716$ & $0.390 \pm 0.016$ \\
        SD 3             & $32.192 \pm 0.738$ & $0.276 \pm 0.016$ & $28.876 \pm 0.686$ & $0.338 \pm 0.016$ \\
        SD 3.5           & $31.050 \pm 0.656$ & $0.287 \pm 0.016$ & $30.383 \pm 0.696$ & $0.308 \pm 0.016$ \\
        SD XL            & $26.299 \pm 0.566$ & $0.353 \pm 0.015$ & $26.632 \pm 0.640$ & $0.358 \pm 0.015$ \\
        \bottomrule
    \end{tabular}
\end{table}

\begin{table}[htbp]
    \centering
    \caption{Performance on Emotional concepts, reported as mean $\pm$ 95\% CI half-width.}
    \label{tab_emotional_ci}
    \small
    \setlength{\tabcolsep}{8pt}
    \renewcommand{\arraystretch}{1.15}
    \begin{tabular}{lcccc}
        \toprule
        & \multicolumn{2}{c}{Clipart} & \multicolumn{2}{c}{Natural} \\
        \cmidrule(lr){2-3}
        \cmidrule(lr){4-5}
        Model & EMD $\downarrow$ & PCC $\uparrow$ & EMD $\downarrow$ & PCC $\uparrow$ \\
        \midrule
        Flux             & $27.959 \pm 1.821$ & $0.344 \pm 0.042$ & $28.718 \pm 2.336$ & $0.407 \pm 0.050$ \\
        OmniGen          & $33.320 \pm 1.939$ & $0.290 \pm 0.042$ & $32.201 \pm 2.322$ & $0.342 \pm 0.048$ \\
        OmniGen2         & $35.260 \pm 2.020$ & $0.261 \pm 0.044$ & $30.346 \pm 2.334$ & $0.348 \pm 0.048$ \\
        Pixart-$\alpha$   & $26.070 \pm 1.739$ & $0.425 \pm 0.044$ & $30.032 \pm 2.136$ & $0.389 \pm 0.048$ \\
        Qwen-Image       & $31.981 \pm 1.763$ & $0.287 \pm 0.039$ & $28.129 \pm 2.073$ & $0.381 \pm 0.048$ \\
        Sana             & $25.058 \pm 1.757$ & $0.438 \pm 0.045$ & $28.033 \pm 2.230$ & $0.425 \pm 0.048$ \\
        SD 3             & $30.748 \pm 2.134$ & $0.297 \pm 0.042$ & $27.379 \pm 2.214$ & $0.374 \pm 0.045$ \\
        SD 3.5           & $31.768 \pm 1.949$ & $0.291 \pm 0.041$ & $29.108 \pm 2.100$ & $0.329 \pm 0.044$ \\
        SD XL            & $27.412 \pm 1.496$ & $0.368 \pm 0.035$ & $27.415 \pm 2.159$ & $0.378 \pm 0.047$ \\
        \bottomrule
    \end{tabular}
\end{table}

\par\endgroup
\begingroup
\section{Evaluation of Recent and Closed-Source Models}
\label{app:recent_models}
We randomly sampled base concepts and included their corresponding Original, Visual, and Emotional variants, resulting in a structured subset of 100 concepts. 
Then, we conducted an additional evaluation of four recent models: FLUX.2 Klein 9B base \citep{team2025flux}, Z-Image \citep{team2025zimage}, GPT Image 2 \citep{openai2026gptimage2}, and Nano Banana 2 \citep{raisinghani2026nanobanana2}. 
The results are reported in the Table~\ref{tab:natural_recent_models} and~\ref{tab:clipart_recent_models}. The best and second best results in each column are marked in \textbf{bold} and \underline{underlined}, respectively.

In the Natural style, Flux2.0 Klein is particularly competitive on Emotional concepts, achieving the best EMD (26.921) and second-best PCC (0.424), while Z-Image obtains the highest Emotional PCC (0.461). In the Clipart style, Nanobanana 2 performs consistently well: it achieves the best Original EMD (27.467), the second-best Visual EMD (28.786), and ranks among the top three on Emotional. GPT Image 2 also obtains the joint-best Visual PCC (0.366) and a strong Emotional PCC (0.435). Overall, the new models improve the leading results in several settings, but no single model dominates across styles and concept types.

\begin{table*}[t]
\centering
\caption{Performance under the Natural style.}
\label{tab:natural_recent_models}
\resizebox{\textwidth}{!}{
\begin{tabular}{lcccccc}
\toprule
Model
& Original EMD $\downarrow$
& Original PCC $\uparrow$
& Visual EMD $\downarrow$
& Visual PCC $\uparrow$
& Emotional EMD $\downarrow$
& Emotional PCC $\uparrow$ \\
\midrule
Flux.1-dev
& \textbf{29.555} & \textbf{0.366}
& 28.114 & 0.373
& 29.460 & 0.344 \\

OmniGen
& 30.838 & 0.338
& 35.046 & 0.305
& 36.085 & 0.271 \\

OmniGen2
& 31.710 & \underline{0.349}
& 34.527 & 0.298
& 32.105 & 0.309 \\

PixArt-$\alpha$
& 31.221 & 0.323
& 30.543 & 0.369
& 36.185 & 0.276 \\

Qwen-Image
& 32.058 & 0.301
& 31.458 & 0.319
& \underline{27.408} & 0.356 \\

SD3
& 29.923 & 0.307
& 31.112 & 0.321
& 27.750 & 0.397 \\

SD3.5
& 31.733 & 0.302
& 34.875 & 0.257
& 31.435 & 0.307 \\

SDXL
& \underline{29.774} & 0.306
& \textbf{26.576} & \underline{0.381}
& 28.389 & 0.374 \\

Sana-1.5
& 30.295 & 0.324
& \underline{27.423} & \textbf{0.423}
& 30.018 & 0.365 \\

Nano Banana 2
& 33.681 & 0.243
& 29.454 & 0.356
& 31.360 & 0.344 \\

GPT Image 2
& 33.091 & 0.263
& 29.968 & 0.357
& 34.485 & 0.349 \\

Flux2.0 Klein 9B Base
& 31.778 & 0.282
& 28.729 & \underline{0.381}
& \textbf{26.921} & \underline{0.424} \\

Z-Image
& 31.970 & 0.307
& 30.459 & 0.345
& 27.434 & \textbf{0.461} \\
\bottomrule
\end{tabular}
}
\end{table*}

\begin{table*}[t]
\centering
\caption{Performance under the Clipart style.}
\label{tab:clipart_recent_models}
\resizebox{\textwidth}{!}{
\begin{tabular}{lcccccc}
\toprule
Model
& Original EMD $\downarrow$
& Original PCC $\uparrow$
& Visual EMD $\downarrow$
& Visual PCC $\uparrow$
& Emotional EMD $\downarrow$
& Emotional PCC $\uparrow$ \\
\midrule
Flux
& 28.585 & 0.367
& 30.890 & 0.322
& 25.585 & 0.289 \\

Omni
& 31.214 & \underline{0.383}
& 37.470 & 0.265
& 33.650 & 0.345 \\

Omni2
& 37.022 & 0.346
& 42.367 & 0.218
& 30.005 & 0.420 \\

PixArt
& 30.998 & 0.324
& 35.207 & 0.304
& \textbf{23.207} & \textbf{0.499} \\

Qwen
& 32.026 & 0.298
& 33.052 & 0.246
& 27.744 & 0.311 \\

SD3
& 30.401 & 0.330
& 35.218 & 0.258
& 27.069 & 0.345 \\

SD3.5
& 32.289 & 0.283
& 33.170 & 0.315
& 29.070 & 0.312 \\

SDXL
& \underline{27.660} & 0.347
& \textbf{27.731} & \underline{0.354}
& 28.089 & 0.318 \\

Sana-1.5
& 28.670 & \textbf{0.391}
& 31.269 & \textbf{0.366}
& \underline{23.527} & \underline{0.472} \\

Nano Banana 2
& \textbf{27.467} & \underline{0.383}
& \underline{28.786} & 0.345
& 23.851 & 0.452 \\

GPT Image 2
& 32.503 & 0.366
& 30.924 & \textbf{0.366}
& 29.471 & 0.435 \\

Flux2.0 Klein 9B Base
& 30.519 & 0.327
& 30.016 & 0.342
& 24.882 & 0.409 \\

Z-Image
& 38.669 & 0.226
& 45.123 & 0.183
& 35.037 & 0.231 \\
\bottomrule
\end{tabular}
}
\end{table*}

\newpage

\par\endgroup
\onecolumn
{
\scriptsize
\begin{longtable}{|l|l|l|p{6cm}|p{4cm}|} 
\caption{Concept list}
\label{tab:condept_list}
\\ \hline
\textbf{Id} & \textbf{Category} & \textbf{Concept} & \textbf{Visual state modifier} & \textbf{Emotional modifier} \\
\hline
\endfirsthead

\hline
\textbf{Id} & \textbf{Category} & \textbf{Concept} & \textbf{Visual state modifier} & \textbf{Emotional modifier} \\
\hline
\endhead

\hline
\endfoot

\hline
\endlastfoot

1 & animal & dog &  &  \\
2 & animal & bird &  &  \\
3 & animal & horse &  &  \\
4 & animal & chicken &  &  \\
5 & animal & bear &  &  \\
6 & animal & cat &  &  \\
7 & animal & wolf &  &  \\
8 & animal & deer &  &  \\
9 & animal & bull &  &  \\
10 & animal & cow &  &  \\
11 & animal & lion &  &  \\
12 & animal & duck &  &  \\
13 & animal & pig &  &  \\
14 & animal & bat &  &  \\
15 & animal & whale &  &  \\
16 & animal & seal &  &  \\
17 & animal & bee &  &  \\
18 & animal & sheep &  &  \\
19 & animal & elephant &  &  \\
20 & animal & shark &  &  \\
21 & animal & rabbit &  &  \\
22 & animal & monkey &  &  \\
23 & animal & goat &  &  \\
24 & animal & butterfly &  &  \\
25 & animal & crab &  &  \\
26 & animal & frog &  &  \\
27 & animal & turtle &  &  \\
28 & animal & crow &  &  \\
29 & animal & goose &  &  \\
30 & animal & spider &  &  \\
31 & animal & ant &  &  \\
32 & animal & dolphin &  &  \\
33 & animal & lobster &  &  \\
34 & animal & owl &  &  \\
35 & animal & coral &  &  \\
36 & animal & squirrel &  &  \\
37 & animal & camel &  &  \\
38 & animal & pigeon &  &  \\
39 & animal & swan &  &  \\
40 & animal & donkey &  &  \\
41 & fruit & apple & fresh, unripe, overripe, rotten, juiceless &  \\
42 & fruit & apricot & fresh, unripe, overripe, rotten, juiceless &  \\
43 & fruit & avocado & fresh, unripe, overripe, rotten, juiceless &  \\
44 & fruit & banana & fresh, unripe, overripe, rotten, juiceless &  \\
45 & fruit & blackberry & fresh, unripe, overripe, rotten, juiceless &  \\
46 & fruit & blueberry & fresh, unripe, overripe, rotten, juiceless &  \\
47 & fruit & cantaloupe & fresh, unripe, overripe, rotten, juiceless &  \\
48 & fruit & cherry & fresh, unripe, overripe, rotten, juiceless &  \\
49 & fruit & coconut & fresh, unripe, overripe, rotten, juiceless &  \\
50 & fruit & cranberry & fresh, unripe, overripe, rotten, juiceless &  \\
51 & fruit & dragonfruit & fresh, unripe, overripe, rotten, juiceless &  \\
52 & fruit & durian & fresh, unripe, overripe, rotten, juiceless &  \\
53 & fruit & fig & fresh, unripe, overripe, rotten, juiceless &  \\
54 & fruit & grape & fresh, unripe, overripe, rotten, juiceless &  \\
55 & fruit & grapefruit & fresh, unripe, overripe, rotten, juiceless &  \\
56 & fruit & kiwi & fresh, unripe, overripe, rotten, juiceless &  \\
57 & fruit & lemon & fresh, unripe, overripe, rotten, juiceless &  \\
58 & fruit & lime & fresh, unripe, overripe, rotten, juiceless &  \\
59 & fruit & lychee & fresh, unripe, overripe, rotten, juiceless &  \\
60 & fruit & mango & fresh, unripe, overripe, rotten, juiceless &  \\
61 & fruit & melon & fresh, unripe, overripe, rotten, juiceless &  \\
62 & fruit & mulberry & fresh, unripe, overripe, rotten, juiceless &  \\
63 & fruit & olive & fresh, unripe, overripe, rotten, juiceless &  \\
64 & fruit & orange & fresh, unripe, overripe, rotten, juiceless &  \\
65 & fruit & papaya & fresh, unripe, overripe, rotten, juiceless &  \\
66 & fruit & peach & fresh, unripe, overripe, rotten, juiceless &  \\
67 & fruit & pear & fresh, unripe, overripe, rotten, juiceless &  \\
68 & fruit & pineapple & fresh, unripe, overripe, rotten, juiceless &  \\
69 & fruit & plum & fresh, unripe, overripe, rotten, juiceless &  \\
70 & fruit & pomegranate & fresh, unripe, overripe, rotten, juiceless &  \\
71 & fruit & pomelo & fresh, unripe, overripe, rotten, juiceless &  \\
72 & fruit & raspberry & fresh, unripe, overripe, rotten, juiceless &  \\
73 & fruit & star fruit & fresh, unripe, overripe, rotten, juiceless &  \\
74 & fruit & strawberry & fresh, unripe, overripe, rotten, juiceless &  \\
75 & fruit & watermelon & fresh, unripe, overripe, rotten, juiceless &  \\
76 & food & bagel & browned, buttery, charred, chocolaty, cinnamon, nutty, stale &  \\
77 & food & biscuit & browned, buttery, charred, chocolaty, cinnamon, nutty, stale &  \\
78 & food & bread & browned, buttery, charred, chocolaty, cinnamon, nutty, stale, toasted &  \\
79 & food & cookie & browned, buttery, charred, chocolaty, cinnamon, cooled, nutty, stale &  \\
80 & food & nut & browned, buttery, charred, cinnamon, stale, toasted &  \\
81 & food & pancake & browned, buttery, charred, chocolaty, cinnamon, nutty, stale, vanilla &  \\
82 & food & sandwich & browned, charred, chocolaty, cinnamon, stale &  \\
83 & food & toast & browned, buttery, charred, chocolaty, cinnamon, nutty, stale &  \\
84 & food & beer & buttery, cooled, stale &  \\
85 & food & brownie & buttery, stale &  \\
86 & food & cereal & buttery, charred, chocolaty, stale &  \\
87 & food & dumpling & buttery, charred, stale &  \\
88 & food & hamburger & buttery, charred, stale &  \\
89 & food & muffin & buttery, chocolaty, nutty, stale, vanilla &  \\
90 & food & oatmeal & buttery, chocolaty, cinnamon, nutty, stale, vanilla &  \\
91 & food & pie & buttery, charred, chocolaty, cinnamon, stale &  \\
92 & food & pudding & buttery, chocolaty, cinnamon, nutty, stale, vanilla &  \\
93 & food & curry & charred, stale &  \\
94 & food & egg & charred, stale, toasted &  \\
95 & food & pasta & charred, stale &  \\
96 & food & pizza & charred, stale &  \\
97 & food & steak & charred, stale &  \\
98 & food & butter & chocolaty, stale &  \\
99 & food & candy & chocolaty, nutty, stale &  \\
100 & food & milk & chocolaty, cinnamon, nutty, stale, vanilla &  \\
101 & food & yogurt & chocolaty, cinnamon, cooled, nutty, stale, vanilla &  \\
102 & food & coffee & cinnamon, cooled, nutty, stale, vanilla, buttery &  \\
103 & food & tea & cinnamon, stale &  \\
104 & food & jelly & cooled, stale &  \\
105 & food & juice & cooled, stale &  \\
106 & food & water & cooled &  \\
107 & food & chocolate & nutty, stale, vanilla &  \\
108 & food & birthday cake & stale &  \\
109 & food & cheese & stale &  \\
110 & food & honey & stale &  \\
111 & food & rice & stale &  \\
112 & food & salad & stale &  \\
113 & food & sushi & stale &  \\
114 & food & jam &  &  \\
115 & food & noodle &  &  \\
116 & plant & acorn & shrivelled, fresh &  \\
117 & plant & aloe & shrivelled, fresh &  \\
118 & plant & bamboo & shrivelled, fresh &  \\
119 & plant & basil & shrivelled, fresh &  \\
120 & plant & bush & shrivelled, fresh &  \\
121 & plant & cactus & shrivelled, fresh &  \\
122 & plant & carnation & shrivelled, fresh &  \\
123 & plant & cherry blossom & shrivelled, fresh &  \\
124 & plant & clover & shrivelled, fresh &  \\
125 & plant & daisy & shrivelled, fresh &  \\
126 & plant & dandelion & shrivelled, fresh &  \\
127 & plant & ginkgo & shrivelled, fresh &  \\
128 & plant & ivy & shrivelled, fresh &  \\
129 & plant & jasmine & shrivelled, fresh &  \\
130 & plant & lavender & shrivelled, fresh &  \\
131 & plant & lily & shrivelled, fresh &  \\
132 & plant & lotus & shrivelled, fresh &  \\
133 & plant & maple & shrivelled, fresh &  \\
134 & plant & marigold & shrivelled, fresh &  \\
135 & plant & mint & shrivelled, fresh &  \\
136 & plant & moss & shrivelled, fresh &  \\
137 & plant & oak & shrivelled, fresh &  \\
138 & plant & orchid & shrivelled, fresh &  \\
139 & plant & palm tree & shrivelled, fresh &  \\
140 & plant & peony & shrivelled, fresh &  \\
141 & plant & pine tree & shrivelled, fresh &  \\
142 & plant & rose & shrivelled, fresh &  \\
143 & plant & seaweed & shrivelled, fresh &  \\
144 & plant & straw & shrivelled, fresh &  \\
145 & plant & sunflower & shrivelled, fresh &  \\
146 & plant & tulip & shrivelled, fresh &  \\
147 & plant & violet & shrivelled, fresh &  \\
148 & plant & wheat & shrivelled, fresh &  \\
149 & vegetables & artichoke & fresh, shrivelled, rotten, juiceless &  \\
150 & vegetables & arugula & fresh, shrivelled, rotten, juiceless &  \\
151 & vegetables & asparagus & fresh, shrivelled, rotten, juiceless &  \\
152 & vegetables & bean & fresh, shrivelled, rotten, juiceless &  \\
153 & vegetables & bell pepper & fresh, shrivelled, rotten, juiceless &  \\
154 & vegetables & bok choy & fresh, shrivelled, rotten, juiceless &  \\
155 & vegetables & broccoli & fresh, shrivelled, rotten, juiceless &  \\
156 & vegetables & brussels sprouts & fresh, shrivelled, rotten, juiceless &  \\
157 & vegetables & cabbage & fresh, shrivelled, rotten, juiceless &  \\
158 & vegetables & carrot & fresh, shrivelled, rotten, juiceless &  \\
159 & vegetables & cauliflower & fresh, shrivelled, rotten, juiceless &  \\
160 & vegetables & celery & fresh, shrivelled, rotten, juiceless &  \\
161 & vegetables & chive & fresh, shrivelled, rotten, juiceless &  \\
162 & vegetables & corn & fresh, shrivelled, rotten, juiceless &  \\
163 & vegetables & cucumber & fresh, shrivelled, rotten, juiceless &  \\
164 & vegetables & eggplant & fresh, shrivelled, rotten, juiceless &  \\
165 & vegetables & garlic & fresh, shrivelled, rotten, juiceless &  \\
166 & vegetables & gourd & fresh, shrivelled, rotten, juiceless &  \\
167 & vegetables & jalapeno & fresh, shrivelled, rotten, juiceless &  \\
168 & vegetables & kale & fresh, shrivelled, rotten, juiceless &  \\
169 & vegetables & leek & fresh, shrivelled, rotten, juiceless &  \\
170 & vegetables & lettuce & fresh, shrivelled, rotten, juiceless &  \\
171 & vegetables & okra & fresh, shrivelled, rotten, juiceless &  \\
172 & vegetables & onion & fresh, shrivelled, rotten, juiceless &  \\
173 & vegetables & parsley & fresh, shrivelled, rotten, juiceless &  \\
174 & vegetables & pea & fresh, shrivelled, rotten, juiceless &  \\
175 & vegetables & pepper & fresh, shrivelled, rotten, juiceless &  \\
176 & vegetables & pickle & fresh, shrivelled, rotten, juiceless &  \\
177 & vegetables & potato & fresh, shrivelled, rotten, juiceless &  \\
178 & vegetables & pumpkin & fresh, shrivelled, rotten, juiceless &  \\
179 & vegetables & radish & fresh, shrivelled, rotten, juiceless &  \\
180 & vegetables & scallion & fresh, shrivelled, rotten, juiceless &  \\
181 & vegetables & spinach & fresh, shrivelled, rotten, juiceless &  \\
182 & vegetables & sprouts & fresh, shrivelled, rotten, juiceless &  \\
183 & vegetables & squash & fresh, shrivelled, rotten, juiceless &  \\
184 & vegetables & sweet potato & fresh, shrivelled, rotten, juiceless &  \\
185 & vegetables & tomato & fresh, shrivelled, rotten, juiceless &  \\
186 & vegetables & zucchini & fresh, shrivelled, rotten, juiceless &  \\
187 & landscape & beach & arid, autumnal, brooding, desolate, dirty, fiery, magical, misty, moonlit, polluted, rainy, spring, stormy, summery, sunny, tropical, vibrant & lonely, oppressive, serene, terrifying, warm \\
188 & landscape & cave & arid, autumnal, brooding, desolate, dirty, forbidding, grassy, magical, misty, muddy, mysterious & lonely, oppressive, serene, somber, terrifying \\
189 & landscape & desert & arid, barren, brooding, desolate, fiery, forbidding, magical, misty, moonlit, mysterious, polluted, rainy, spectacular, spring, summery & oppressive, serene, terrifying \\
190 & landscape & forest & arid, autumnal, barren, brooding, dense, desolate, ethereal, fiery, forbidding, lush, magical, misty, moonlit, muddy, mysterious, rainy, spring, stormy, summery, sunny, tropical, verdant, vibrant & cozy, lonely, oppressive, serene, terrifying, warm \\
191 & landscape & grassland & arid, autumnal, barren, brooding, desolate, forbidding, magical, misty, moonlit, muddy, mysterious, olive, rainy, spring, stormy, summery, sunny, tropical, verdant, vibrant & cozy, lonely, oppressive, serene, warm \\
192 & landscape & lake & arid, autumnal, blushing, brooding, clear, dirty, ethereal, forbidding, frozen, magical, misty, moonlit, murky, mysterious, polluted, rainy, spring, stormy, summery, sunny, tropical, turbulent, vibrant & lonely, oppressive, serene, terrifying, warm \\
193 & landscape & mountain & arid, autumnal, barren, brooding, desolate, forbidding, grassy, magical, majestic, misty, moonlit, muddy, mysterious, olive, rainy, spectacular, spring, stormy, summery, sunny, tropical, verdant, vibrant & lonely, oppressive, serene, terrifying \\
194 & landscape & island & autumnal, barren, brooding, desolate, dirty, dreamy, forbidding, grassy, magical, misty, moonlit, muddy, mysterious, polluted, rainy, spring, stormy, summery, tropical, vibrant & lonely, oppressive, serene, terrifying \\
195 & landscape & sunrise & autumnal, blushing, fiery, magical, misty, spectacular, spring, summery, tropical & cozy, hopeful, oppressive, serene, terrifying, warm \\
196 & landscape & sunset & autumnal, blushing, fiery, magical, misty, spectacular, spring, summery, tropical & cozy, oppressive, serene, terrifying, warm \\
197 & landscape & moon & blushing, brooding, forbidding, magical, mysterious, pale, summery & oppressive, somber, terrifying, wistful \\
198 & landscape & sky & blushing, brooding, clear, cloudless, cloudy, faded, fiery, forbidding, magical, misty, murky, mysterious, pale, polluted, rainy, spring, starry, stormy, sunny, thunderous, tropical & cozy, hopeful, oppressive, serene, somber, terrifying, wistful \\
199 & landscape & sun & blushing, fiery, magical, summery, tropical & hopeful, oppressive, terrifying, warm \\
200 & landscape & glacier & brooding, forbidding, magical, majestic, misty, mysterious, polluted, rainy, spectacular, spring & oppressive, serene \\
201 & landscape & ocean & brooding, clear, desolate, dirty, forbidding, frozen, magical, misty, moonlit, murky, mysterious, polluted, rainy, spectacular, spring, stormy, summery, sunny, tropical, turbulent & oppressive, serene, terrifying, warm \\
202 & landscape & waterfall & brooding, dirty, ethereal, forbidding, frozen, magical, majestic, misty, moonlit, murky, mysterious, polluted, spectacular, spring, stormy, turbulent & oppressive, terrifying \\
203 & landscape & rainbow & magical & hopeful \\
204 & building & cabin & ancient, autumnal, brooding, dirty, forbidding, grassy, magical, mysterious, stormy, stylish, vintage & somber, terrifying, warm, oppressive, serene, cozy, lonely \\
205 & building & castle & ancient, autumnal, brooding, desolate, dirty, dreamy, forbidding, grassy, magical, majestic, mysterious, solemn, spectacular, stormy, stylish, sumptuous, vintage & somber, terrifying, oppressive, serene, lonely \\
206 & building & church & ancient, autumnal, brooding, desolate, dirty, dreamy, forbidding, grassy, magical, majestic, mysterious, solemn, spectacular, stormy, stylish, sumptuous, vintage & somber, terrifying, oppressive, serene, lonely \\
207 & building & hospital & ancient, brooding, dirty, forbidding, magical, mysterious & somber, terrifying, oppressive, serene \\
208 & building & palace & ancient, autumnal, brooding, desolate, dirty, dreamy, forbidding, grassy, magical, majestic, mysterious, solemn, spectacular, stormy, stylish, sumptuous, vintage & somber, terrifying, oppressive, serene, lonely \\
209 & building & pyramid & ancient, autumnal, brooding, desolate, forbidding, magical, majestic, mysterious, solemn, spectacular, stormy & somber, terrifying, oppressive, serene, lonely \\
210 & building & temple & ancient, autumnal, brooding, desolate, dirty, fiery, forbidding, grassy, magical, majestic, mysterious, solemn, spectacular, stormy & somber, terrifying, oppressive, serene, lonely \\
211 & building & farm & arid, autumnal, barren, brooding, desolate, dirty, forbidding, magical, misty, mysterious, rainy, spring, stormy, summery, tropical, verdant, vibrant & hopeful, terrifying, serene, oppressive, cozy \\
212 & building & aquarium & brooding, dirty, dreamy, forbidding, magical, mysterious, spectacular, sumptuous & somber, oppressive, serene \\
213 & building & casino & brooding, dreamy, forbidding, magical, mysterious, spectacular, sumptuous, vintage & somber, oppressive, serene \\
214 & building & factory & brooding, desolate, dirty, forbidding, magical, mysterious, spectacular, vintage & somber, terrifying, oppressive, serene, lonely \\
215 & building & igloo & brooding, desolate, forbidding, magical, mysterious & somber, oppressive, serene, lonely \\
216 & building & skyscraper & brooding, forbidding, magical, majestic, mysterious, spectacular, stylish, sumptuous & somber, terrifying, oppressive, serene, lonely \\

\end{longtable}
}

\end{document}